\documentclass[acmtog, dvipsnames, screen=true, authorversion]{acmart} %

\usepackage{tabularx,booktabs} %
\usepackage[normalem]{ulem}  %
\usepackage{color,hyphenat,balance}
\usepackage[fleqn,tbtags]{mathtools}
\usepackage{overpic}  %
\usepackage{wrapfig}
\usepackage{contour}
\usepackage{microtype}
\usepackage{multirow}
\usepackage{colortbl}
\usepackage{amsmath}
\usepackage{bbm}
\usepackage{textcomp}
\usepackage{enumitem}
\setitemize{noitemsep,topsep=0pt,parsep=0pt,partopsep=0pt}
\usepackage{derivative}
\usepackage{float}

\usepackage{listings}
\usepackage{graphicx}
\graphicspath{{./images/}}
\usepackage[ruled]{algorithm2e} 

\SetAlFnt{\small}
\SetAlCapFnt{\small}
\SetAlCapNameFnt{\small}
\SetAlCapHSkip{0pt}

\def \eg {{\emph{e.g.},\thinspace}}
\def \ie {{\emph{i.e.},\thinspace}}

\newcommand{\rev}[1]{{\color{black} #1}}
\newcommand{\mypar}[1]{\vspace{1mm}\noindent\textbf{#1}}

\definecolor{llightgray}{RGB}{230,230,230}

\newcommand{\name}{\emph{Sketch2Arti}\xspace}
\newcommand{\dname}{\emph{SketchMobility}\xspace}

\setcopyright{cc}
\setcctype{by-nc-nd}
\acmJournal{TOG}
\acmYear{2026} \acmVolume{45} \acmNumber{4} \acmArticle{108}
\acmMonth{7} \acmDOI{10.1145/3811375}

\begin{document}

\title{\name: Sketch-based Articulation Modeling of CAD Objects}

\author{Yi Yang}
\orcid{https://orcid.org/0009-0007-4814-1733}
\affiliation{%
  \institution{University of Edinburgh}
  \streetaddress{10 Crichton Street}
  \city{Edinburgh}
  \country{United Kingdom}
}
\email{arloyang397@gmail.com}

\author{Hao Pan}
\orcid{https://orcid.org/0000-0003-3628-9777}
\affiliation{%
 \institution{Tsinghua University}
 \streetaddress{}
 \city{Beijing}
 \country{China}
 }
 \email{hp.panhao@hotmail.com}

\author{Yijing Cui}
\orcid{0009-0006-7647-604X}
\affiliation{%
 \institution{Tsinghua University}
 \streetaddress{}
 \city{Beijing}
 \country{China}
 }
 \email{yijingcui2003@gmail.com}
 
\author{Alla Sheffer}
\orcid{https://orcid.org/0000-0001-9251-3716}
\affiliation{%
 \institution{University of British Columbia}
 \streetaddress{}
 \city{Vancouver}
 \country{Canada}
 }
\email{sheffa@cs.ubc.ca}

\author{Changjian Li}
\orcid{0000-0003-0448-4957}
\affiliation{%
 \institution{University of Edinburgh}
 \streetaddress{10 Crichton Street}
 \city{Edinburgh}
 \country{United Kingdom}
 }
 \email{chjili2011@gmail.com}

\begin{teaserfigure}
\centering
\begin{overpic}[width=0.99\linewidth]{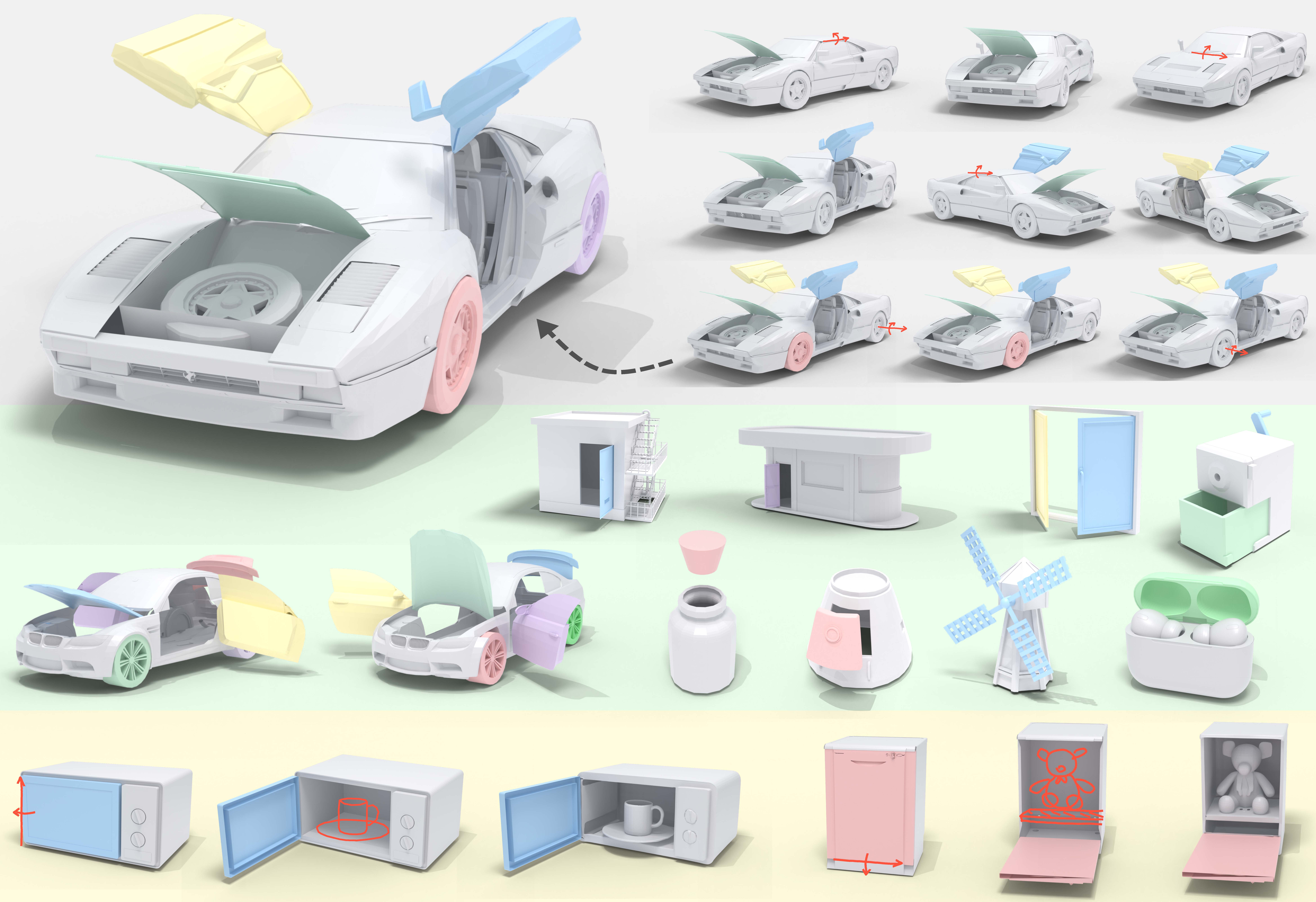}
\end{overpic}
\caption{\textbf{Sketch-based articulation modeling.}
We present \name, the first sketch-based system for articulation modeling of CAD objects. 
\name is versatile. 
\emph{Top}: through iterative sketch-based editing, \name progressively discovers multiple movable parts and recovers their motion parameters on a complex car model.
\emph{Middle-left:} \name offers high controllability---e.g., a car door can be opened in different user-specified manners (standard, backward-hinged, or butterfly) by simply changing the sketches.
\emph{Middle-right:} \name generalizes to diverse objects beyond public datasets, enabling articulation modeling for unseen categories.
\emph{Bottom:} for shell models lacking internal structures, users can sketch the missing components \rev{(\eg the teddy bear)} and \name faithfully generates plausible internal mechanisms that respect both the existing geometry and the predicted articulation parameters.
}
\label{fig:teaser}
\end{teaserfigure}

\begin{abstract}

Articulation modeling aims to infer movable parts and their motion parameters for a 3D object, enabling interactive animation, simulation, and shape editing.
In this paper, we present \name, the first \emph{sketch-based articulation modeling} system for CAD objects.
Our key observation is that designers naturally communicate articulation intent through lightweight sketches (\eg arrows and strokes) that indicate how parts should move, yet translating such sketches into articulated 3D models remains largely manual. 
\name bridges this gap by enabling users to specify articulation through simple 2D sketches drawn from a chosen viewpoint. 
Given a CAD model and user sketches, our approach automatically discovers the corresponding movable parts and predicts their motion parameters, allowing iterative modeling of multiple articulations on complex objects with fine-grained control.
Importantly, \name is trained in a category-agnostic manner without requiring object category information, leading to strong generalization to diverse objects beyond existing articulation datasets.
Moreover, for shell models lacking interior structures, \name supports controllable internal completion guided by user sketches, generating plausible internal components consistent with the existing geometry and predicted motion constraints.
Comprehensive experiments and user evaluations demonstrate the effectiveness, controllability, and generalization of \name.

\emph{The code, dataset, and the prototype system are at \rev{\url{https://arlo-yang.github.io/Sketch2Arti}}.}

\end{abstract}



\maketitle

\section{Introduction}

Articulated objects are ubiquitous in everyday environments.
From furniture and tools to vehicles and mechanical devices, articulation enables objects to function as intended and supports safe and intuitive human-object interaction.
Modeling articulation, \ie identifying movable parts and recovering their motion parameters, is therefore a fundamental problem in computer graphics, with broad applications in animation, simulation, and interactive shape editing.

Recent research has made encouraging progress on articulation modeling from visual and textual inputs, including texts~\cite{su2025artformer}, images~\cite{chen2025freeart3d,yuan2025larm}, videos~\cite{le2024articulate,lu2025generating, vora2025articulate}, and point clouds~\cite{jiang2022ditto}.
These works address articulation perception~\cite{hu2017learning,sharf2014mobility}, articulated objects reconstruction~\cite{jiang2022ditto,Liu_2023_ICCV}, and articulated objects generation~\cite{liu2024singapo,chen2025freeart3d}, often leveraging learned priors from the PartNet-Mobility dataset~\cite{xiang2020sapien} and its variants.
However, existing datasets cover only limited categories and shape instances, which inevitably restricts model capabilities beyond benchmark settings.
As a result, current methods typically offer limited controllability, generalize poorly to out-of-distribution objects, and exhibit strong data-specific biases, making them difficult to deploy in practical design workflows.

\begin{figure}[!t]
    \centering
    \includegraphics[width=\linewidth]{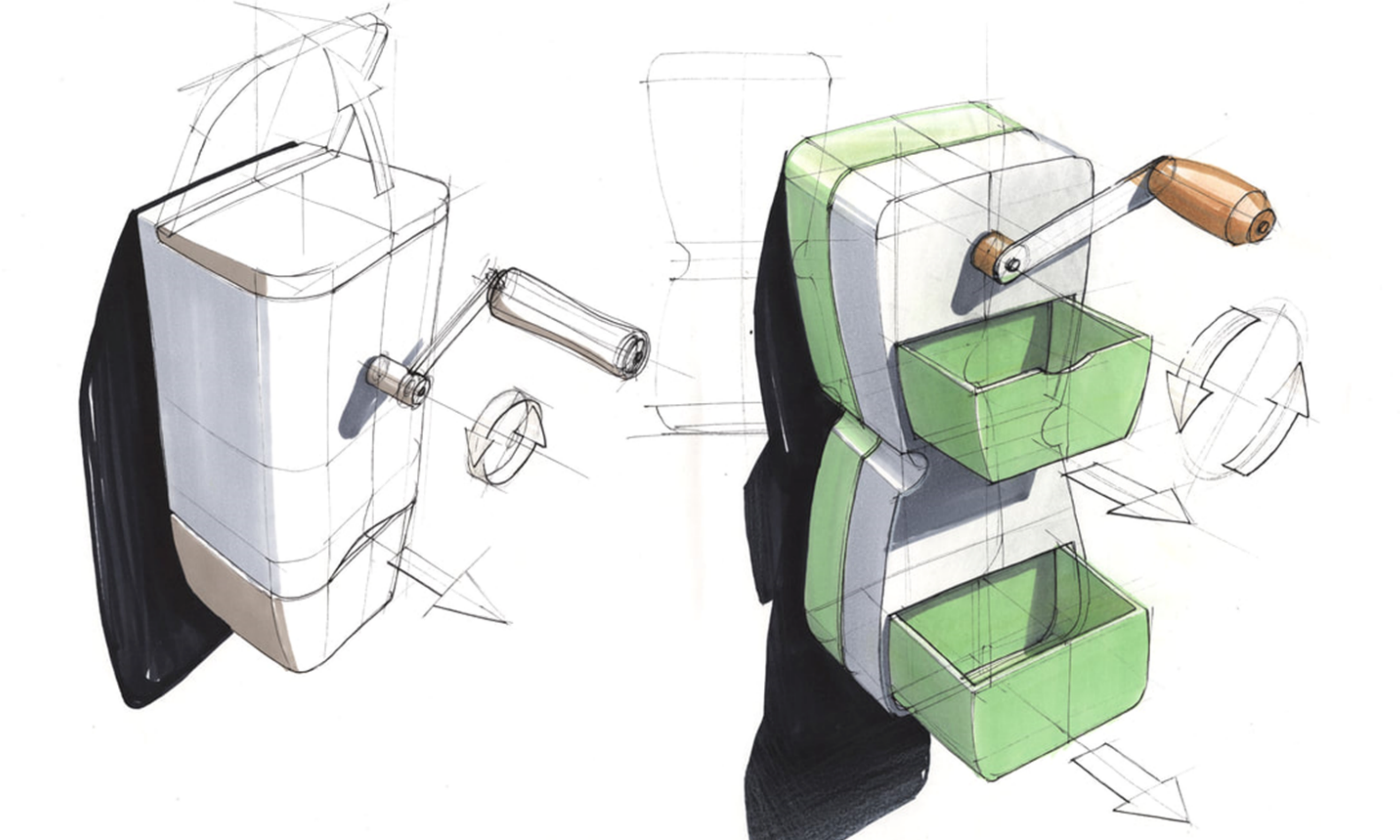}
    \caption{\textbf{Articulation modeling in the design field.} 
    In the product design workflow, designers frequently draw arrow-like strokes depicting the articulation cues of man-made objects in the ideation stage. Other than the arrows, strokes representing the part after articulation (lid of the left container) and unseen internal structure (the drawer of the right container) are drawn to express the geometry.
    The design sketches are usually translated \emph{manually} in the professional software, lacking advanced tools for automatic conversion.
    Sketch artwork \textcopyright \emph{Susie Brand-de Groot}, used with permission.
    }
    \label{fig:motivation}
\end{figure}

On the other hand, in the mechanical design practice, articulation design as an inherent step of CAD modeling has a succinct but expressive form in the ideation stage, where sketches are used for describing both shape and articulation \cite{eissen2008sketching,eissen2019sketching}. 
Figure~\ref{fig:motivation} demonstrates typical ideation sketches in real design workflows.
Designers naturally express articulation intent using lightweight sketches, such as arrows or the anticipated end poses after articulation (see the drawer).
These sketches clearly convey both how parts should move and which geometric components are involved.
Nevertheless, to the best of our knowledge, there exists no technique to turn such sketches into articulations of digital 3D assets, calling for an automatic solution with high controllability and strong generalization.

Motivated by this observation, we revisit articulated modeling and propose the new task: \emph{sketch-based articulation modeling}. 
Given a 3D shape and a few user strokes indicating articulation intent, our goal is to identify the movable parts and infer their motion parameters by jointly analyzing user strokes and input geometry.
This task poses several challenges, including i) determining \emph{where} the articulated part is located,  ii) inferring \emph{how} it moves (\eg joint type and motion parameters), and iii) optionally completing \emph{what} is occluded and becomes visible under articulation, especially when the input is a shell model lacking interior structure.

We propose \name, addressing these challenges through two tightly coupled components: (1) sketch-guided articulation analysis and prediction, and (2) articulation-aware controllable generation that respects both the existing geometry and the inferred motion constraints.
By design, \name provides strong user controllability via lightweight interaction.
For example, users can specify different articulation intents for the same object part, such as opening a microwave door from left-to-right, right-to-left, or top-to-bottom, simply by changing the sketches.
To achieve strong generalization, our predictor deliberately avoids relying on global object category semantics.
Instead, it only observes local geometric cues \rev{(\eg depth and normal maps)} around the sketched region, enabling component-level articulation inference that generalize across categories.
We further adopt a multi-task learning formulation that jointly predicts movable-part cues and motion parameter.
Given the predicted part cues, we extract geometry- and feature-preserving movable parts using a foundation segmentation network (\ie Partfield~\cite{liu2025partfield}), producing clean articulated components and completing the \emph{where} and \emph{how} inference.

Beyond articulation prediction, we additionally tackle the \emph{what} challenge: completing or creating geometry revealed by articulation.
This is particularly important for shell meshes (\eg shapes from generative models~\cite{zhang2024clay}), where internal structures are missing and become exposed after articulation.
Moreover, even for complete CAD objects, users may wish to interactively model internal content, such as placing a cup inside an opened microwave, to support downstream animation.
To this end, \name supports sketch-guided completion and insertion while maintaining geometric consistency with the input shape.
The key challenge lies in generating new geometry that both preserves the existing structure and strictly respects the inferred articulation parameters.
We address this by adapting a state-of-the-art 3D generative model---Trellis~\cite{xiang2025structured}, into an articulation-aware shape inpainting module, and incorporate a geometry-based motion-parameter correction procedure.
Intuitively, applying the predicted articulation should not introduce physical conflicts (\eg a closed drawer should not intersect the cabinet body).
Our correction module enforces such geometric validity during generation, yielding faithful completions while simultaneously refining motion parameters.

To train and evaluate sketch-based articulation modeling, we introduce a new dataset---\dname, containing $\sim$4.7K articulated shape instances across diverse categories.
For each shape, we curate or refine articulation annotations (including both existing parameters and newly labeled ones), and synthesize diverse user sketches that reflect realistic articulation intent.
Each instance thus provides the original geometry, movable-part labels, motion parameters, and corresponding sketch inputs.
We will release the dataset to foster future research on controllable, sketch-driven articulation modeling.

We conduct extensive experiments to validate the effectiveness, controllability, and generalization of \name.
Our method consistently outperforms SoTA baselines across diverse settings, and ablation studies further justify key design choices.
We additionally integrate our algorithm into a prototype interactive system and perform a user study with novice users, demonstrating that \name enables intuitive and efficient articulation modeling in practice.

To summarize, our principal contributions are four-fold:
\begin{itemize}
    \item We introduce \name, the first sketch-based articulation modeling system for CAD objects, enabling intuitive and fine-grained user control for articulation creation.
    \item We propose a category-agnostic articulation prediction framework that leverages sketch-local geometric cues and jointly infers movable parts and motion parameters, leading to strong generalization beyond PartNet-Mobility.
    \item We adapt 3D generative models for articulation-aware, sketch-controllable shape inpainting, enabling completion of missing internal structures while respecting the inferred motion constraints.
    \item We present \dname, a benchmark dataset with articulated CAD models and synthetic sketch annotations, facilitating training and benchmarking for sketch-driven articulation modeling.
\end{itemize}

\section{Related Work}

Our work aims to bridge the gap between sketch-based modeling and articulation modeling.
For a broader overview, we refer interested readers to the surveys on  
sketch-based modeling~\cite{bonnici2019sketch,tono2026deep} and CAD articulation modeling~\cite{liu2025survey}, which cover a wide range of relevant literature.
We next discuss the prior works most closely related to our work.

\paragraph{Articulation modeling.}
Recent progress in 3D content creation has strengthened the priors available for object geometry (and often appearance), making high-quality \emph{static} assets increasingly accessible. This includes diffusion-guided text-to-3D optimization pipelines~\cite{poole2022dreamfusion} as well as native 3D generative models trained at scale that decode to common 3D formats~\cite{zhang2024clay,xiang2025structured}. With these foundations, a natural next step for functional assets is to explicitly model \emph{articulation}: how an object decomposes into parts and the constraints governing their motion.

Articulation has been studied extensively under articulation understanding \rev{\cite{xu2009joint, mitra2010illustrating, goyal2025geopard}}, where methods infer movable parts and joint parameters (e.g., joint type, axis, pivot, and state) from observations.
In addition, recent work injects higher-level semantic priors via vision-language models to support articulation reasoning and produce structured outputs suitable for downstream interaction~\cite{Huang2024A3VLM}.

Beyond understanding, articulated reconstruction \rev{\cite{song2024reacto, mandi2024real2code, weng2024neural, jiang2022ditto}} aims to recover 3D shape together with kinematic structure. A prominent family of approaches builds on differentiable rendering and neural object representations. NeRF-style optimization~\cite{mildenhall2021nerf} and 3D Gaussian Splatting~\cite{kerbl20233d} offer convenient backbones for fitting geometry/appearance to images, and have been extended to articulated settings by introducing part-wise motion models and category-level priors. Related work also learns class-level priors over articulated shape and motion from images, providing regularization when observations are sparse or ambiguous~\cite{wei2022self}. \rev{More recently, \cite{yuan2025larm} trains a feedforward model to synthesize multiview multi-state images to guide explicit articulated object reconstruction.}

Generative articulation modeling \rev{\cite{gao2025meshart,li2024dragapart, le2024articulate}} aims to learn distributions over articulated objects, enabling the creation of novel instances with plausible part structure and motion constraints, optionally conditioned on images or other inputs. Neural 3D Articulation Prior (NAP) introduces a generative formulation that jointly models articulated shape and kinematic structure~\cite{lei2023nap}. Subsequent work such as SINGAPO~\cite{liu2024singapo} constructs articulated household objects from a single image by predicting part connectivity and kinematic attributes and assembling retrieved part geometry. 
\cite{su2025artformer} applies transformer model to generate articulated objects under tree structure constraints.
\cite{chen2025freeart3d} adapts an existing static 3D generative model \cite{xiang2025structured} for dynamic articulation generation via score distillation sampling over multiple image-specified states given as input.

\rev{Despite the variety of approaches, a common bottleneck is the heavy reliance on constrained datasets like PartNet-Mobility \cite{xiang2020sapien}. Beyond its limited scale and category coverage, previous works often further restrict their training and evaluation to specific object subsets, leading to poor generalization to novel geometries.}
In this paper, we demonstrate that with a minimum amount of sketching as input, and via the new approach of sketch-based articulation modeling, significant improvements in generalization, quality and controllability can be achieved for articulation modeling of static CAD objects (Sec.~\ref{subsec:comp}).

\paragraph{Sketch-based articulation modeling.}

While sketch based modeling has been extensively studied for shape creation and editing \cite{bonnici2019sketch}, few works have explored the use of sketches for modeling motions and articulations.
\cite{Lee_2022_Rapid} showcases an interactive sketch-based system for simultaneous modeling of 3D curve networks and their articulation; however, the shapes in the form of rough curves are not directly usable in animation and simulation.
\cite{ChoiSketchiMo2016} applies sketching to animate articulated characters that have already been rigged.
To our best knowledge, our work for the first time proposes to use sketching to specify articulations for generic 3D CAD models without predefined structural decomposition into movable parts.

\begin{figure}[!t]
    \centering
    \begin{overpic}[width=\linewidth]{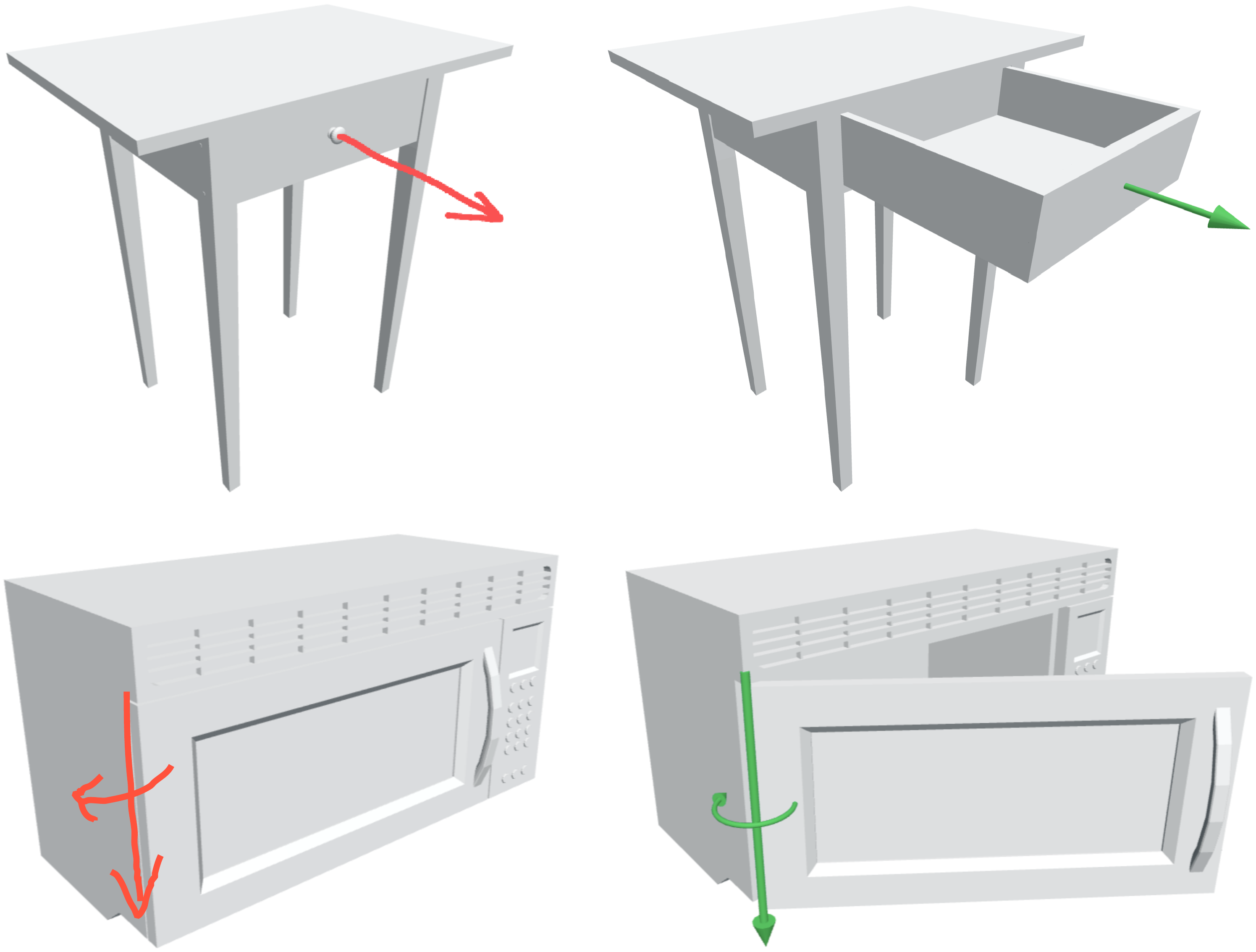} %
        \put(15, -3.5) {\small (a) User sketch}
        \put(60, -3.5) {\small (b) Motion parameters}
        \put(98,55) {\small $a_t$}
        \put(92, 61.5) {\small $d_t$}
        \put(61,22) {\small $p_r$}
        \put(62, 1) {\small $a_r$}
        \put(53, 11) {\small $\theta_r$}
    \end{overpic}
    \caption{\textbf{Task input and articulation parameterization.} (a) Users draw sketches (red) over the object image. (b) The parameters of translation (top) and rotation (bottom) are shown with green arrows and overlaid symbols. Note that the pivot point $p_r$ could be any point along the side edge of the door. 
    }
    \label{fig:arti_parameters}
\end{figure}

\begin{figure}[!t]
    \centering
    \begin{overpic}[width=\linewidth]{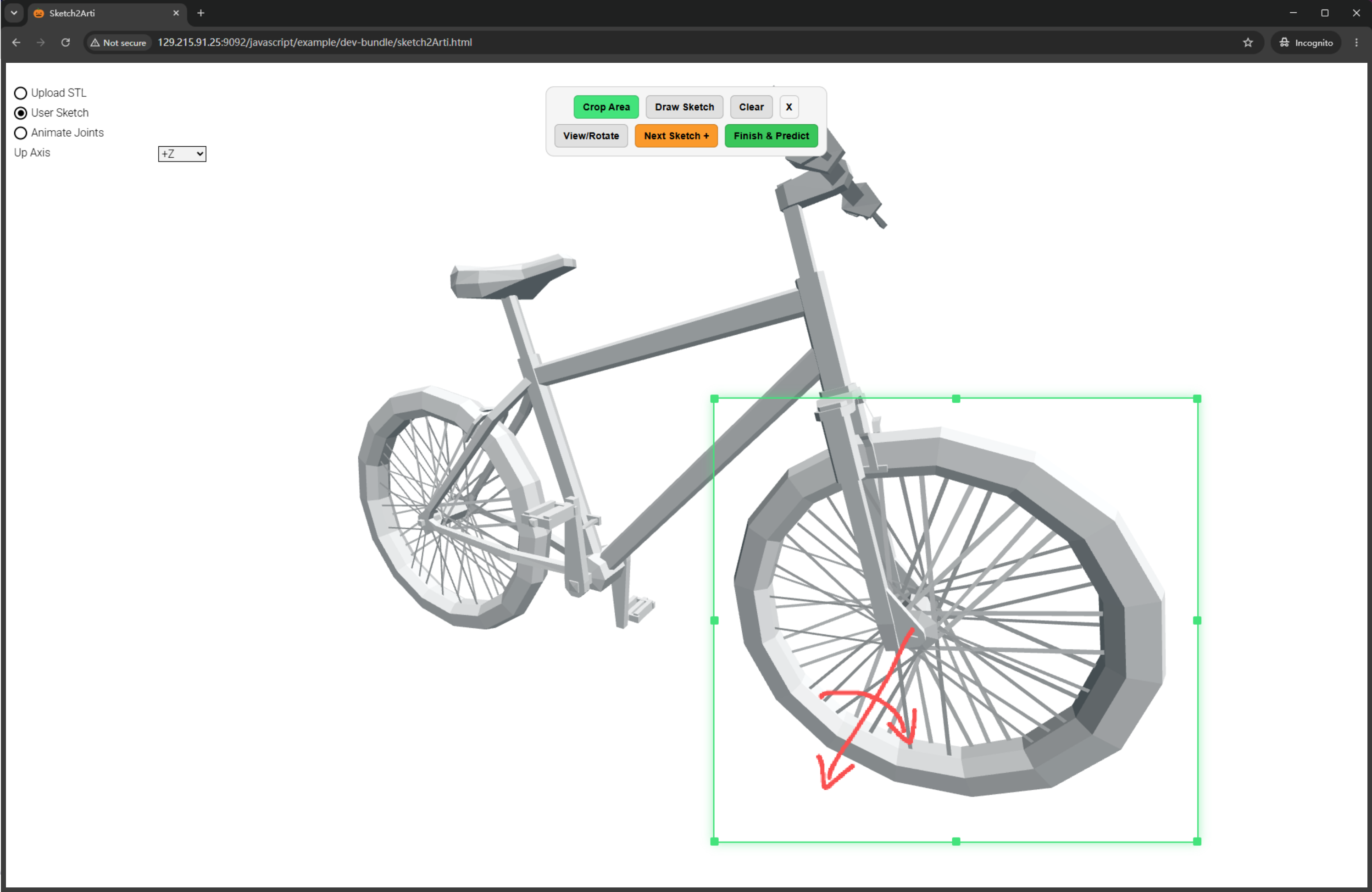} %
    \end{overpic}
    \caption{\textbf{User interface.} The interface consists of an operation menu (top left, \eg  load an object), an interaction menu (top right, \eg draw sketch), and a wide user interaction panel. 
    After loading the object, users freely choose the desired view, select a focal field (green), draw strokes (red) indicating articulation intention, and click the ``Finish \& Predict'' button to obtain the result. 
    The green box shows the perceptive focal field used to locate local shape context.
    }
    \label{fig:user_interface}
\end{figure}

\begin{figure*}[!t]
    \centering
    \begin{overpic}[width=\textwidth]{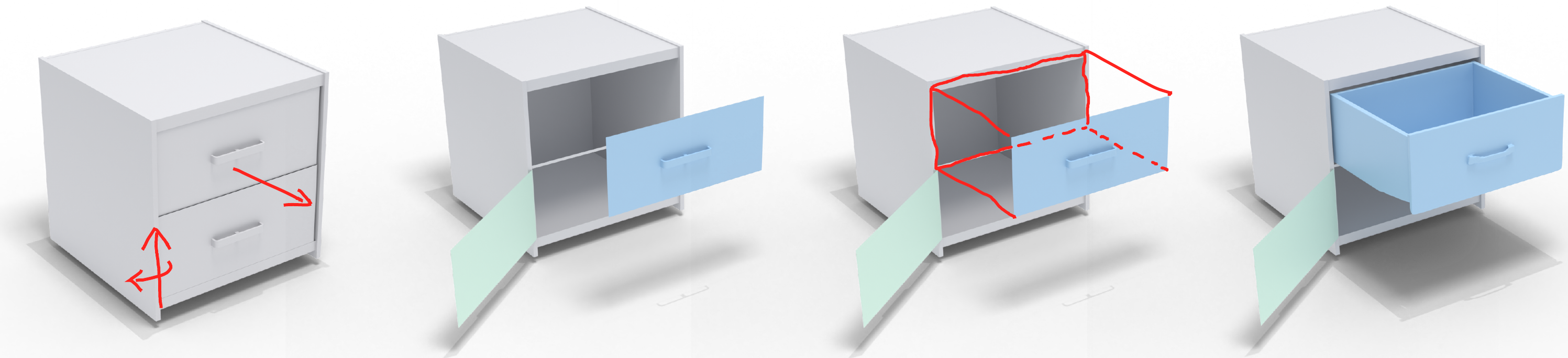} %
        \put(10,-1.5) {\small (a) Input}
        \put(32,-1.5) {\small (b) Articulated object}
        \put(57,-1.5) {\small (c) Interior sketching}
        \put(83,-1.5) {\small (d) Completed object}
    \end{overpic}
    \caption{\textbf{Overview.} (a) Given an input 3D shape and the user sketches, our method \name addresses the \emph{where} and \emph{how} challenges by (b) identifying movable parts (\ie the two doors) and inferring their articulation parameters. (c) The predicted motion reveals missing internal structure (\eg an empty drawer), which users can further specify via sketches. \name then tackles the \emph{what} challenge by (d) generating the full drawer geometry while adhering to both the existing shape and the inferred articulation.
    }
    \label{fig:overview}
\end{figure*}
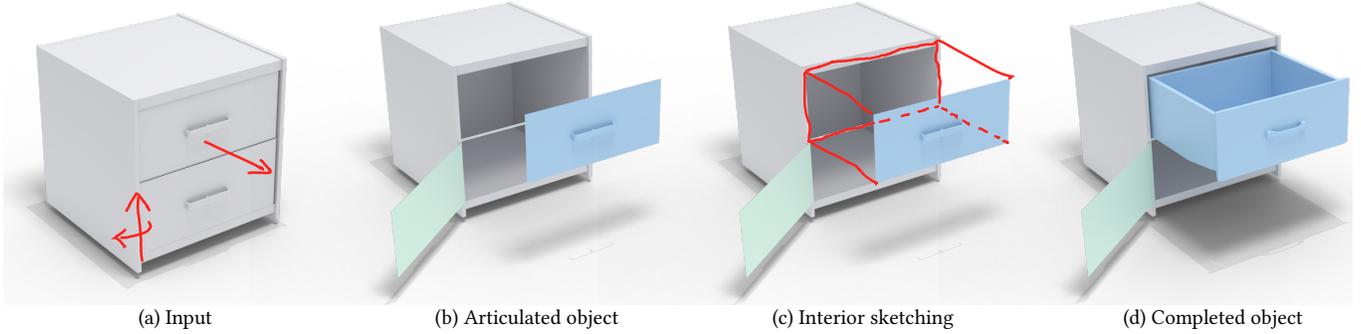

\section{Overview}
\label{sec:overview}

We introduce the task of \emph{sketch-based articulation modeling}, which aims to identify the movable part(s) of a 3D shape and infer their articulation parameters by jointly analyzing the user-provided sketch and the underlying 3D geometry.

\mypar{Input.}
The input consists of a 3D object $O$ (with or without texture) represented as a mesh model, a user-selected viewpoint $v$, and the user-drawn sketch $S$. 
Specifically, the object is viewed under $v$ to show an image $\pi_v(O)$, over which the user draws a sketch $S$ that specifies articulation intent (see Fig.~\ref{fig:arti_parameters}(a)).

\mypar{Output.}
The output is a movable object part $o \in O$ and its articulation, parameterized by its motion type $t \in \{\texttt{rotation}, \texttt{translation}\}$ and the corresponding motion parameters. \rev{Specifically, a rotational articulation is characterized by} a hinge axis defined by a pivot point $p_r\in \mathbb{R}^3$ and a unit-length direction vector $a_r\in \mathbb{R}^3$, together with the range of rotation $[0,\theta_r]$, 
\rev{while a translational articulation is defined by} a unit translation direction $a_t \in \mathbb{R}^3$, and the range of translation $[0,d_t]$.
Fig.~\ref{fig:arti_parameters}(b) displays the respective parameterization.
Note that in this paper, we focus on two articulation types, \ie rotation and translation, as they cover the majority of articulated motions in everyday objects.
We do not predict the motion range \rev{in our system}, since it is highly under-constrained from sketch input alone and often depends on user preference and design intent.
Instead, we assign a reasonable default range for visualization (\eg \rev{100}$^\circ$ for a door) and allow users to \emph{interactively} adjust the range of motion after articulation modeling.
This design keeps the interaction lightweight while preserving user control over the final articulation. %

\mypar{User interface.}
Our system employs lightweight user interaction designed to bridge the gap between human intent and 3D articulation (see Fig.~\ref{fig:user_interface}). 
After loading an arbitrary 3D object, users select a preferred viewpoint and sketch directly on the rendered object to indicate the desired motion. Specifically, translation is specified by a single directional arrow, whereas rotation is defined by a hinge axis line paired with a directional arrow indicating the opening direction. 
These 2D user cues are then lifted into a geometry-aware representation, which drives our articulation prediction and completion. 
Importantly, \name naturally supports iterative articulation modeling: after modeling the articulation of one part, users can interactively navigate to another viewpoint and continue sketching to specify additional movable parts, enabling progressive articulation authoring for complex objects with multiple degrees of freedom.

\mypar{Method.}
Figure~\ref{fig:overview} shows a high-level overview of our method and also the key challenges of the task: \emph{where} the movable part is, \emph{how} it moves, and \emph{what} internal geometry should be completed or revealed under articulation. To address the challenges, we have proposed several novel solutions:
\begin{itemize}
    \item \emph{Where}: given a localized region containing the target part and a sketch depicting its motion, we deduce the part mask by first predicting a 2D mask on $\pi_v(O)$,  roughly designating the part projection, based on which \rev{we} then selecting the most fitting 3D part via the help of a hierarchical 3D segmentation foundation model.
    
    \item \emph{How}: along with the 3D part segmentation, we predict the motion parameters from the joint analysis of the sketch and the underlying geometry. The estimated articulation parameters may not be accurate, and we refine them to ensure geometric validity and physical plausibility. 
    In particular, we snap the rotation pivot to meaningful geometric anchors (\eg part boundaries or centers) and align rotation/translation axes with local surface normals or principal directions, preventing collisions and misalignment.
    
    \item \emph{What}: when the input shape is incomplete (e.g., shell meshes) or when users request additional internal structures, we rely on generative models to complete the interior structures. 
    Any adapted generative model needs to complete what's missing and preserve what's given, although most generative models have been trained extensively on the exterior surfaces mostly.
    We develop masked and iterative generation to solve this constrained completion problem.
\end{itemize}

\begin{figure*}[!t]
    \centering
    \begin{overpic}[width=\textwidth]{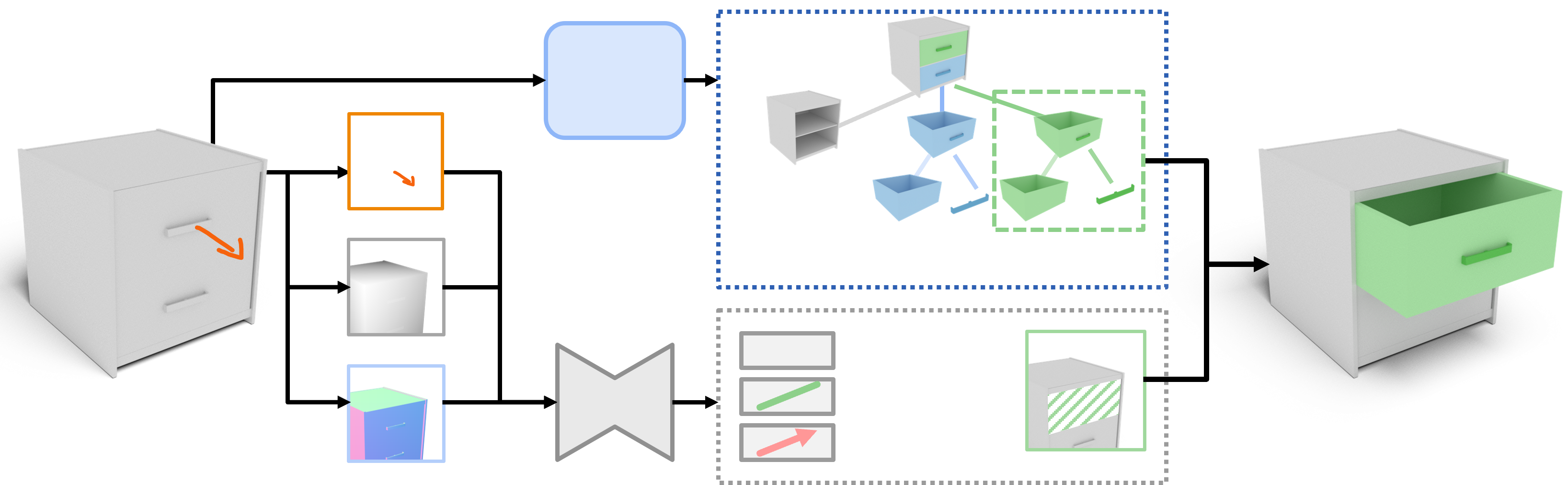} %
    \put(3, 5) {\small Shape and Sketching}
    
    \put(23.4,16.3) {\small \rev{sketch}}
    \put(23.4,8.3) {\small depth}
    \put(23.2,0.1) {\small normal}

    \put(36.4,25.2) {Partfield}
    \put(37.3,4.8) {U-Net}

    \put(49.0,8.2) {\small R / T}
    \put(54.9,8.2) {\small Motion Type}
    \put(54.9,5.2) {\small Pivot Map}
    \put(54.9,2.2) {\small Direction}
    \put(48.5,13.5) {\small \rev{Agglomerative Tree Merging Cluster}}
    
    \put(66.3,0.8) {\small Mask map}

    \put(84, 4.5) {\small Articulated Shape}
    
    \end{overpic}
    \caption{\textbf{Articulation prediction}. Given a static 3D object, we apply category-agnostic articulation recognition on a localized region surrounding the sketch with the local context captured by the depth and normal maps. A trained U-Net module predicts the articulation parameters in 2D maps and 3D local camera coordinates, as well as motion type. The 2D part mask is then back-projected onto the object surface and used to filter through a hierarchy of segments produced by a foundation Partfield model, to select the best matching part at a level undetermined beforehand as the movable 3D component.
    }
    \label{fig:arti_prediction}
\end{figure*}
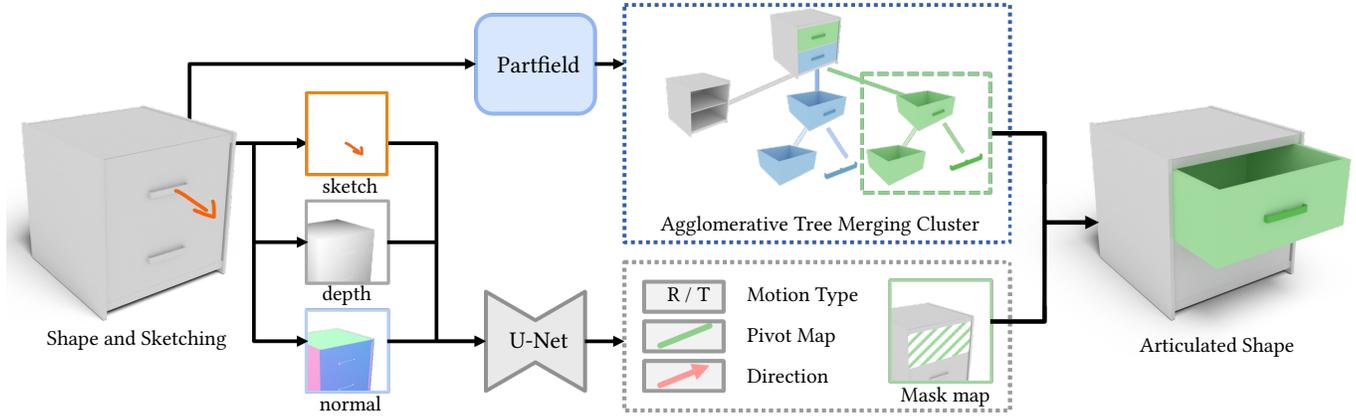

\section{Sketch-based Articulation Analysis}
\label{sec:articulation_analysis}

Given rough sketches depicting the articulation intention of unspecified part of a 3D object, we propose a hybrid approach containing robust neural network-based recognition and subsequent matching and refinement for precision, to reconstruct the intended part (\textit{where}) and its articulation parameters (\textit{how}).
Figure~\ref{fig:arti_prediction} illustrates the steps of this stage.

\subsection{Network-based Articulation Recognition}
\label{subsec:arti_recognition}

Neural networks provide robust solutions for recognizing rough sketches \cite{bonnici2019sketch}. 
In particular, we draw on inspirations of \cite{li2020sketch2cad,li2022free2cad} to achieve generalizable recognition of 2D sketches, so that \name can be applied to novel object categories.

The idea of generalizable recognition is to limit the input for neural network to a localized region surrounding the sketch, such that the network learns to recognize local movable parts and estimate their motion parameters from a local context, independent of the whole object which the part belongs to.
As such, we construct the input consisting of a 5-channel map depicting the local region, including a single channel sketch, a 3-channel surface normal map, and a single channel depth map that are concatenated.
This rich input representation allows the model to correlate the sparse sketch of 2D semantic guidance with both local surface orientation and absolute spatial geometry.

We develop a lightweight CNN to process the input data.
To predict the articulation parameters, we employ a UNet-based architecture that operates on this image-space representation. 
The network is designed with a multi-head decoder to simultaneously predict four kinematic attributes: (1) a segmentation mask identifying the movable pixels; (2) a pivot heatmap representing the probability distribution of the joint axis's 2D projection; (3) the joint type (classification logits for \rev{translation vs. rotation}); and (4) the 3D motion direction (a normalized vector in camera space). 
Given these attributes, we can robustly recover the full kinematic configuration: the 2D pivot and mask provide spatial localization (including the pivot $p_r$ and rough region for the movable part $o$) by back-projecting onto the 3D model, while the direction head directly lifts the motion vector into $a_r$ or $a_t$ in the 3D space.

\rev{To facilitate \emph{generalizable} network learning, our network features two unique technical designs.
Firstly, unlike prior methods that depend on category-specific shape priors for articulation estimation \cite{chen2025freeart3d,liu2024singapo}, our model utilizes the user sketch as a direct and explicit motion cue. 
Secondly, the output of 2D maps and estimated orientations in local camera coordinates align well with the input maps.}
This shifts the learning focus towards local geometric features, enabling the network to resolve articulation based on local functional geometries—such as cylindrical knobs for rotation or planar panels for sliding—that are shared across diverse categories.
In Sec.~\ref{subsec:comp} we show the improved generalization of our localized articulation recognition.

\subsection{Movable Part Segmentation}
\label{subsec:segmentation}

Given the 2D part mask, we map it to the 3D movable part by leveraging a 3D foundation model---Partfield~\cite{liu2025partfield}.
Partfield provides a feature field that specifies how each point over the object surface is different from the other points, and has shown great generalization by training on large-scale data and under the 2D foundation model supervision.

Nonetheless, 
the naive clustering (\eg by k-means) on Partfield features produces segmentations that not necessarily match with articulation, and it is infeasible to choose from them the proper instance that the recognized motion should be associated with.
To this end, we \rev{follow the agglomerative clustering strategy in Partfield to} apply hierarchical clustering on Partfield features, which produce a tree of segments starting from small details and converging all the way to large components, and use the recognized part mask to filter through the hierarchy of segmentations, to find the best matching part node (\ie one that maximizes the IoU between the mask and the part surface) across the levels of hierarchy as the 3D movable part, as illustrated in Fig.~\ref{fig:arti_prediction}. 
In Sec.~\ref{subsec:ablation_arti}, we ablate the use of filtering such a hierarchical segmentation with a naive k-means clustering and matching, which shows the advantage of our approach.

\subsection{Post-processing}
\label{subsec:post-processing}

The recognized motion parameters, including directions for translation and rotation axis, tend to have numerical errors.
Directly applying them to the selected part will produce intolerable artifacts when the motion range becomes large.
We apply post-processing in the form of snapping to regularize the predicted motion parameters, so that the parts can move following structure constraints. 

\rev{As a preliminary step, for a predicted rotational motion, we further determine if it is continuous (\ie periodic rotation covering the whole range of $[0,2\pi]$) or non-continuous (\ie rotation within a fixed range, as found in an opening door) by examining the local geometry. 
Specifically, we first compute a representative part surface normal (\ie the face area-weighted average of view-facing triangle normals of the movable part), and measure its angular difference with the predicted rotational direction. If the angular difference is smaller than $30^\circ$, we classify the motion as continuous; otherwise, it is considered non-continuous. 
}

\rev{In the following post-processing, for translation and continuous rotation, we 
estimate an oriented bounding box (OBB) of the movable part and snap the direction vector to the closest OBB face normal.}
For example, the car wheels rotate continuously along the direction of the wheel's front surface normal.
For non-continuous rotation, we find the boundary between the movable part and the non-moving structure around the pivot point and align the rotation axis along the boundary tangential direction, with the assumption that such a boundary represents the hinge of the part rotation. 
For example, the fridge door opens along the boundary edge between the door and the body. 

In Sec.~\ref{subsec:ablation_arti}, we ablate the result with and without post-processing, which shows the improved regularity of our approach.

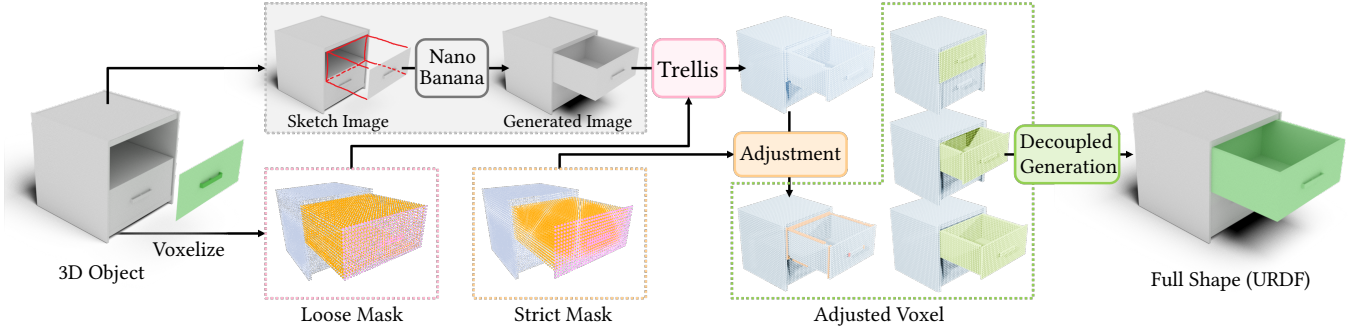
\begin{figure*}[!t]
    \centering
    \begin{overpic}[width=\textwidth]{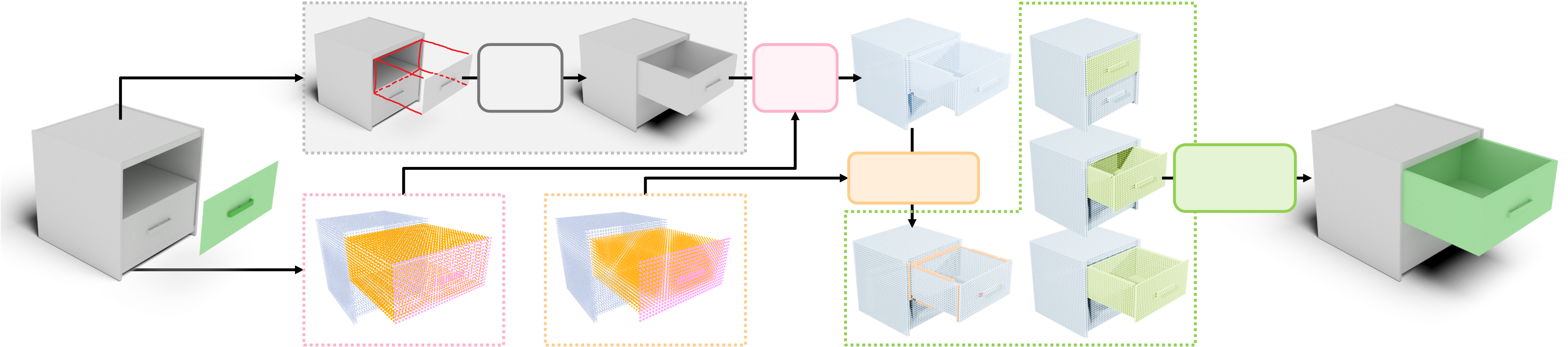} %
    \put(4, 1.5) {\small 3D Object}
    \put(11,3.2) {\small Voxelize}

    \put(21,12.8) {\footnotesize Sketch Image}
    \put(37,12.8) {\footnotesize Generated Image}
    \put(31.5,17.5) {\small Nano}
    \put(30.8,16) {\small Banana}
    \put(22,-1.6) {\small Loose Mask}
    \put(37.8,-1.6) {\small Strict Mask}
    \put(48.4,16.5) {Trellis}

    \put(54.5,10.3) {\small Adjustment}
    \put(60,-1.6) {\small Adjusted Voxel}
    \put(75.3,11.2) {\small Decoupled}
    \put(75.3,9.4) {\small Generation}

    \put(85,1) {\small Full Shape (URDF)}
    \end{overpic}
    \caption{\textbf{Interior shape completion}. Our approach leverages 2D and 3D generative models to complete the interior structures exposed by articulated parts. Given a 3D object with recognized articulation part and parameters, the top branch applies a 2D generative model (\eg Nano banana) to obtain a high-quality reference image, which is used to guide the 3D generative model (\eg Trellis) to create the interior structure. Crucially for obtaining structure-preserving interiors, masks of loose and strict types are built to control the flow generative process of the 3D generative model and adjust the completed part interior, respectively. Finally, the completed part is refined for kinematic validity and turned into separate meshes that are readily usable as URDF models.  
    }
    \label{fig:interior_completion}
\end{figure*}

\section{Controllable Interior Generation}
\label{sec:shape_completion}

For object articulation modeling, in addition to recognizing the motion of the articulated part, a sometimes optional but highly essential requirement is to generate proper interior structures that would become visible after the movable part changed its position.
Prior the existence of 3D generative models, this task is as challenging as modeling a new object from scratch. 
With the emergence of capable generative models, we can leverage them for hallucinating the interior structures.
However, the challenges remain with ensuring the generated interior structure is compatible with both the remaining structures and the articulation motion.
In this section, we present how we have adapted existing 3D generative models to fullfil this articulation-aware constrained generation task.

\subsection{Interior Structure Generation Pipeline}
\label{sec:inpainting_pipeline}

Most existing assets within datasets \cite{deitke2023objaversea,deitke2023objaverseb,wang2025partnext} or synthesized by generative models represent objects as surface-only models, containing only the geometry visible in a canonical static pose. 
Kinematic articulation exposes physical voids in regions that were previously occluded, rendering the object unsuitable for realistic physical interaction.
To resolve this incompleteness, we leverage the robust priors of 2D and 3D generative models, in a pipeline illustrated in Fig.~\ref{fig:interior_completion}. 

We first articulate the object to its open state and generate a completed 2D reference image $I_{\text{comp}}$, with the help of 2D image generative models (\eg Nano banana \cite{GoogleNanoBananaPro2025}). 
This blueprint can be obtained either automatically by inpainting the exposed void using a text-guided 2D diffusion model, or manually via user sketches. This dual-mode approach offers both plausibility and controllability, allowing users to create internal structures absent from 3D dataset priors. 

Next we lift this 2D reference image into a 3D geometry consistent with both the rest shape and given articulation, via a process of image-guided 3D inpainting, as detailed in Sec. \ref{sec:flow_completion}.

\subsection{Structure-Preserving Flow Completion}
\label{sec:flow_completion}

Given the partial 3D geometry $X_{\text{shell}}$ derived from the articulated shell and the reference $I_{\text{comp}}$, our goal is to synthesize the missing geometry $X_{\text{miss}}$. The final output $X_{\text{final}}$ is constrained to satisfy two criteria: (1) \textit{Semantic Consistency}, where the synthesized regions align visually with the structures predicted in $I_{\text{comp}}$; and (2) \textit{Geometric Fidelity}, where the valid regions of the asset remain identical to the input $X_{\text{shell}}$.

To achieve this, we introduce a modified flow matching scheme adapting an existing state-of-the-art 3D generative model Trellis~\cite{xiang2025structured}. 
Trellis encodes a 3D shape as a latent $Z \in \mathbb{R}^{C \times 16^3}$, which is essentially a volumetric grid of resolution $16^3$, with each grid point associated with a latent vector of dimension $C$ encoding the local geometry and appearance.
Furthermore, let \rev{$M_{\mathrm{occ}} \in \{0, 1\}^{C \times 16^3}$} be the binary mask marking the regions occupied by the input shell $X_{\text{shell}}$, and $Z^{shell}$ \rev{as} the shell shape converted into structure latent via Trellis encoder.

\paragraph{Masked Subspace Generation}
As a flow matching model \cite{lipman2023flow}, Trellis applies flow sampling updates to the latent state $Z_t$ at time $t\in [0,1]$ by following the velocity field $v_\theta$ predicted by conditioning on the reference image. 
This yields a predicted next state $Z_{t+\Delta t}^{\text{gen}}$ after $\Delta t$:
\begin{equation}
    Z_{t+\Delta t}^{\text{gen}} = Z_t + \Delta t \cdot v_\theta(Z_t, t \mid I_{\text{comp}}).
\end{equation}

Meanwhile, we analytically derive the reference state for the known shell at the target timestep:
\begin{equation}
    Z_{t+\Delta t}^{\text{shell}} = (t+\Delta t) Z_{\text{shell}} + (1-(t+\Delta t))\epsilon,
\end{equation}
where $\epsilon$ is the sampled initial noise for the flow process. 

To achieve controlled completion, we fuse the generative and analytic states guided by the shell mask \rev{{$M_{\mathrm{occ}}$}}:
\rev{
\begin{equation}
    Z_{t+\Delta t} \leftarrow M_{\mathrm{occ}} \odot Z_{t+\Delta t}^{\mathrm{shell}} + (1-M_{\mathrm{occ}}) \odot Z_{t+\Delta t}^{\mathrm{gen}}.
    \label{eq:fusion-shell-mask}
\end{equation}}
This ensures that at every step of the flow matching integration, the trajectory is anchored to the known geometry.

\paragraph{Spatial Confinement via Negative Masking}
Since the object shell typically occupies a small fraction of the entire volume, the generative model often hallucinates artifacts in the surrounding free space. To prevent this, we introduce a negative masking strategy grounded in the object's articulation kinematics.

Recall that our pipeline identifies the articulation type and trajectory (Sec.~\ref{sec:articulation_analysis}). 
This prior allows us to strictly define the valid generation zone: it is limited to the volume relevant to the motion, such as the sweep path of a sliding drawer or the internal cavity of a cabinet. 
Consequently, we define the "forbidden" mask $M_{\text{void}}$ as the complementary region outside this valid zone. 
The fusion equation (\ref{eq:fusion-shell-mask}) is thus updated as:
\rev{
\begin{equation}
    Z_t \leftarrow M_{\mathrm{occ}} \odot Z_t^{\mathrm{shell}} + M_{\mathrm{void}} \odot Z_t^{\mathrm{empty}} + (1 - M_{\mathrm{occ}} - M_{\mathrm{void}}) \odot Z_t^{\mathrm{gen}},
    \label{eq:fusion-shell-empty-mask}
\end{equation}
}
where $Z_t^{\text{empty}}$ is the empty voxel state encoded by Trellis encoder.

\paragraph{Iterative Growth Strategy}

While the above structure-preserving mechanism guarantees alignment with the input shell, we note that it frequently cannot generate a complete shape in one pass of flow matching sampling (Sec.~\ref{subsec:ablation_completion}).

The reasons can be two-fold. First, existing 3D generative models are extensively trained on shell mesh shapes given by the large scale datasets \cite{deitke2023objaverseb}, and thus has limited capacity in generating interior structures when the movable parts have exposed them.
Second, the constant injection of the partial geometry (Eq.~\ref{eq:fusion-shell-empty-mask}) suppresses the growth of new structures in the boundary regions. 

\begin{figure*}[t]
    \centering
    \begin{minipage}{0.7\textwidth}
    \centering
    \includegraphics[width=\linewidth]{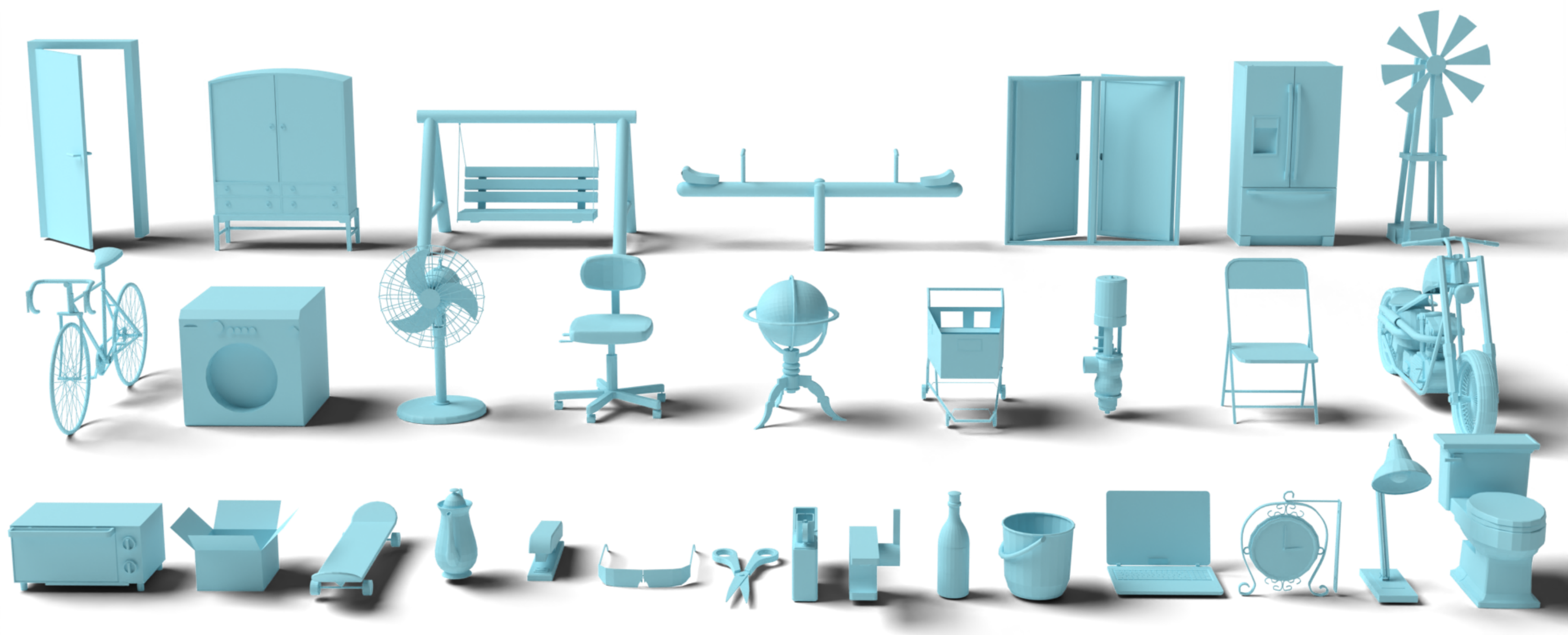}
    \end{minipage}
    \hfill %
    \begin{minipage}{0.28\textwidth}
    \centering
    \includegraphics[width=\linewidth]{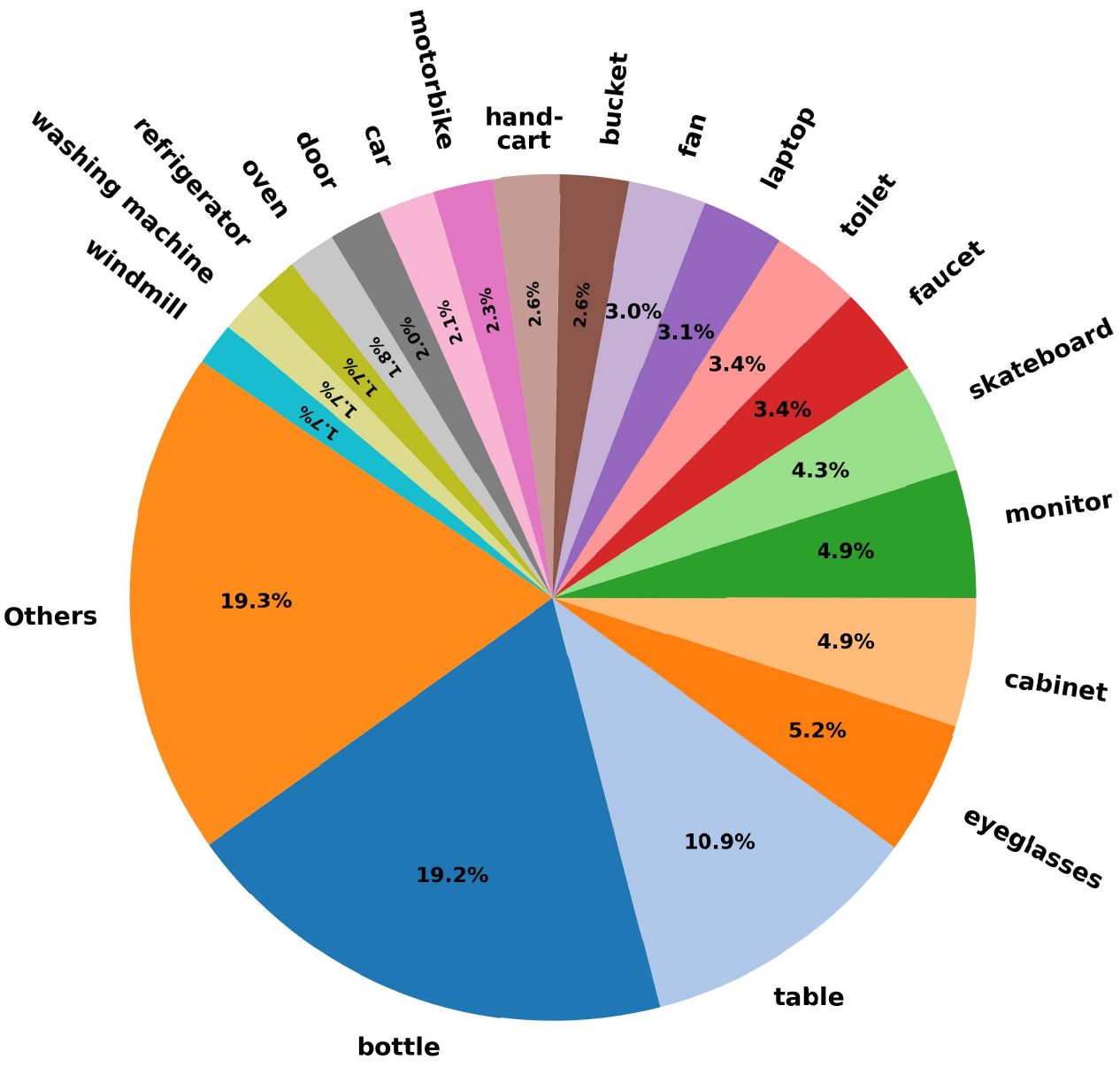}
    \end{minipage}
    \caption{\textbf{Dataset gallery and statistics.}
    \textbf{Left:} Representative samples from \dname. Note the presence of uncommon articulated objects (\eg windmills and motorbikes), which are rarely considered in existing articulation modeling benchmarks.
    \textbf{Right:} Category distribution of \dname. We report major categories ($\geq$1.5\%) individually, while merging minor categories into \textit{Others} (\rev{19.3}\%).
    }
  \label{fig:dataset_gallery}
\end{figure*}

To resolve this, we adopt an iterative refinement strategy. \rev{We let $k$ be the iteration index and $M_{\mathrm{occ}}^{(k)}$ denote the occupancy mask of the known region at step $k$. At initialization, $M_{\mathrm{occ}}^{(0)}$ is given by the occupancy of the input shell $X_{\text{shell}}$.} We treat the output of the current pass as the geometric prior for the next, allowing the structure to stabilize and propagate incrementally. Let $X_{\text{final}}^{(k)}$ be the synthesized geometry from iteration $k$. For iteration $k+1$, we update the structural constraint to include the newly generated content:
\begin{equation}
    X_{\text{shell}}^{(k+1)} \leftarrow X_{\text{final}}^{(k)}.
\end{equation}
This establishes a feedback loop where the model progressively consolidates the hallucinated structure. Empirically, we find that 10--15 iterations are sufficient to resolve complex internal geometries, while simpler cases converge significantly faster (Sec.~\ref{subsec:ablation_completion}).

\subsection{Kinematic Refinement and Decoupling}

To ensure the physical validity of the generated asset, \ie solid parts that can move along the static base shape without collision, we apply strict post-generation refinement grounded in kinematic constraints.

\paragraph{Loose-to-Strict Mask Constraint}
We observe that strictly enforcing the kinematic envelope $1-M_{\text{void}}$ during flow generation can overly constrain the generation. 
To mitigate this, we employ a loose constraint during the iterative generation phase (Sec. \ref{sec:flow_completion}), where the forbidden mask $M_{\text{void}}$ is morphologically eroded by 1 voxel, to propagate geometry more robustly from the known boundary.

Following generation, we apply a strict kinematic filter. We intersect the synthesized voxels with the precise kinematic sweep volume. Any generated geometry falling outside this volume is pruned. This ``generate-then-prune'' strategy balances generative freedom with strict collision avoidance.

\paragraph{Motion Range Calibration}
The generated structures may not fit into the predicted articulation range precisely, such as a drawer generated shorter than its intended slot. 
We perform a post-generation collision analysis along the articulation trajectory. 
By detecting the precise contact points between the synthesized moving part and the static base, we re-calibrate the valid motion range. This ensures that the final asset is not only geometrically clean but also kinematically consistent, strictly respecting the physical boundaries of the generated interior.

\paragraph{Decoupled Mesh Extraction}
A common failure mode in end-to-end 3D generation is the undesirable connection of articulating parts. Since the generated internal structure is spatially contiguous with the static frame, mesh extraction algorithms like FlexiCubes \cite{shen2023flexicubes} often merge them into a single, non-articulated manifold.
To prevent this, we explicitly decouple the processing of dynamic and static components.
\rev{For example, the synthesized internal voxels and the front panel of the drawer in Fig.~\ref{fig:interior_completion} are isolated from the input shell and processed as a separate entity during the mesh extraction phase.  
}
This ensures that the final asset maintains a clean topological separation between moving parts, guaranteeing functional articulation. 

Ultimately, our interior generation pipeline yields a fully disentangled, collision-free, and geometrically completed mesh assembly, which is automatically exported as a URDF model ready for downstream physical simulation.

\section{Building \dname Dataset}

Since no existing dataset supports sketch-based articulation modeling, we construct a new dataset, \dname. 
We describe the dataset construction in detail below.

\paragraph{Data collection.} 
We curated our dataset from three distinct sources to ensure comprehensive coverage of object categories and kinematic structures:

\begin{itemize}
    \item PartNeXt: we utilize PartNeXt \cite{wang2025partnext}, a dataset featuring high-quality models and fine-grained hierarchical segmentations. 
    From its catalog, we focus exclusively on the 15- categories that exhibit inherent articulation, effectively filtering out static object classes. 
    These categories include Monitor, Laptop, Fan, Bucket, Table, Glasses, Pliers, Bottle, Microwave, Oven, Scissors, Skateboard, Toilet, Washing Machine, and Door. 
    After rigorously filtering out instances with suboptimal segmentation, we retain 2,174 validated shapes for kinematic annotation.
    
    \item Shape2Motion: to expand the dataset beyond standard indoor furniture, we incorporate Shape2Motion \cite{wang2019shape2motion}. This source is selected to introduce kinematic diversity through outdoor and mechanical categories, such as Car, Windmill, and Motorbike. We integrate 2,036 shapes across 40 categories, supplementing their high-quality segmentation with precise motion range parameters.

    \item Procedural Generation: recognizing that \rev{cabinet}, dishwasher, microwave, and tables represent a primary use case in articulation modeling tasks, we employ procedural generation~\cite{joshi2505procedural} to synthesize 500 additional instances. This ensures robust structural coverage for these critical categories with pre-defined kinematic chains.
\end{itemize}

In total, \dname comprises 4,710 fully articulated shapes spanning 43 diverse categories. Figure~\ref{fig:dataset_gallery} shows a preliminary gallery (left) and category distribution statistics (right) of the dataset.
Refer to the supplementary for the detailed per-category statistics. 

\paragraph{Data platform and kinematic annotation.} 
To handle these diverse sources, we build a lightweight WebGL-based interface with simple interactions that reduce manual effort and adapt to each source’s needs.
For raw segmentation data like PartNeXt, the annotation involves two key steps. 
First, since the granularity of raw segmentation may not align with the desired articulation definition, we manually click to select and merge relevant fine-grained mesh segments into a single semantic group representing the moving part. 
Second, we assign motion properties to this consolidated group by simply clicking to define the 3D pivot point, specifying the axis orientation, and interactively dragging a control slider to visualize the articulation in real-time, thereby ensuring accurate motion assignment. 
This streamlined efficiency allows annotators to define or refine complex kinematic chains in an average of $1\sim5$ minutes per shape for the manually processed categories.
For Shape2Motion, which already possesses valid part segmentation but lacks physical constraints, our platform is deployed specifically to supplement the missing motion range parameters. 
Finally, for the procedurally generated subset, since these shapes are synthesized with intrinsic kinematic attributes, they bypass the manual annotation phase and are directly prepared for the sketch alignment process described in the subsequent section.

\begin{figure}[!b]
    \centering
    \begin{overpic}[width=\linewidth]{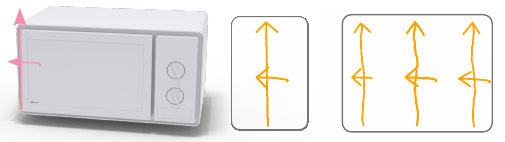} %
    \put(9,-3) {\footnotesize (a) 3D motion cues}
    \put(40,-3) {\footnotesize (b) W/o perturbation}
    \put(67.5,-3) {\footnotesize (c) De-regularized sketches}
    \end{overpic}
    \caption{\textbf{Sketch synthesis.}
    (a) Given a 3D shape and its articulation, we construct 3D motion cues (\eg hinge axis vectors and rotational arcs) to represent the motion of movable parts.
    (b) Directly projecting these 3D cues onto the image plane yields perfectly smooth curves, which are unrealistic for human freehand drawing.
    (c) We therefore inject pixel-level perturbations to obtain synthesized strokes that better resemble human sketches.}
    \label{fig:sketch_synthesis}
\end{figure}

\paragraph{Sketch synthesis.} 
To synthesize sketches that visually communicate the corresponding articulated motion, we implement a pipeline designed to mimic human drawing behavior, and Fig.~\ref{fig:sketch_synthesis} illustrates the process.

Firstly, we extract the ground-truth 3D kinematic trajectory derived from our motion annotations, such as hinge axis vectors and rotational arcs, and map its key coordinates directly into 2D screen space using the camera's projection matrix. 
The camera is placed automatically, facing the movable part to capture the curves clearly.
Rather than simply rendering a rigid 3D arrow model, we discretize this projected path into a sequence of screen-space points. 
To bridge the domain gap between perfect geometric curves and freehand sketching, we apply stochastic geometric perturbations, such as jitter and drift, directly to these 2D points. 
This step simulates the imperfections (especially when drawing with a mouse) and perspective errors inherent in human drawing.
Finally, the perturbed points are fitted with quadratic Bézier curves and rendered onto a canvas to produce the final human-stylized output.
This synthesis improves robustness to diverse user inputs.

\begin{figure*}[!t]
    \centering
    \includegraphics[width=0.99\textwidth]{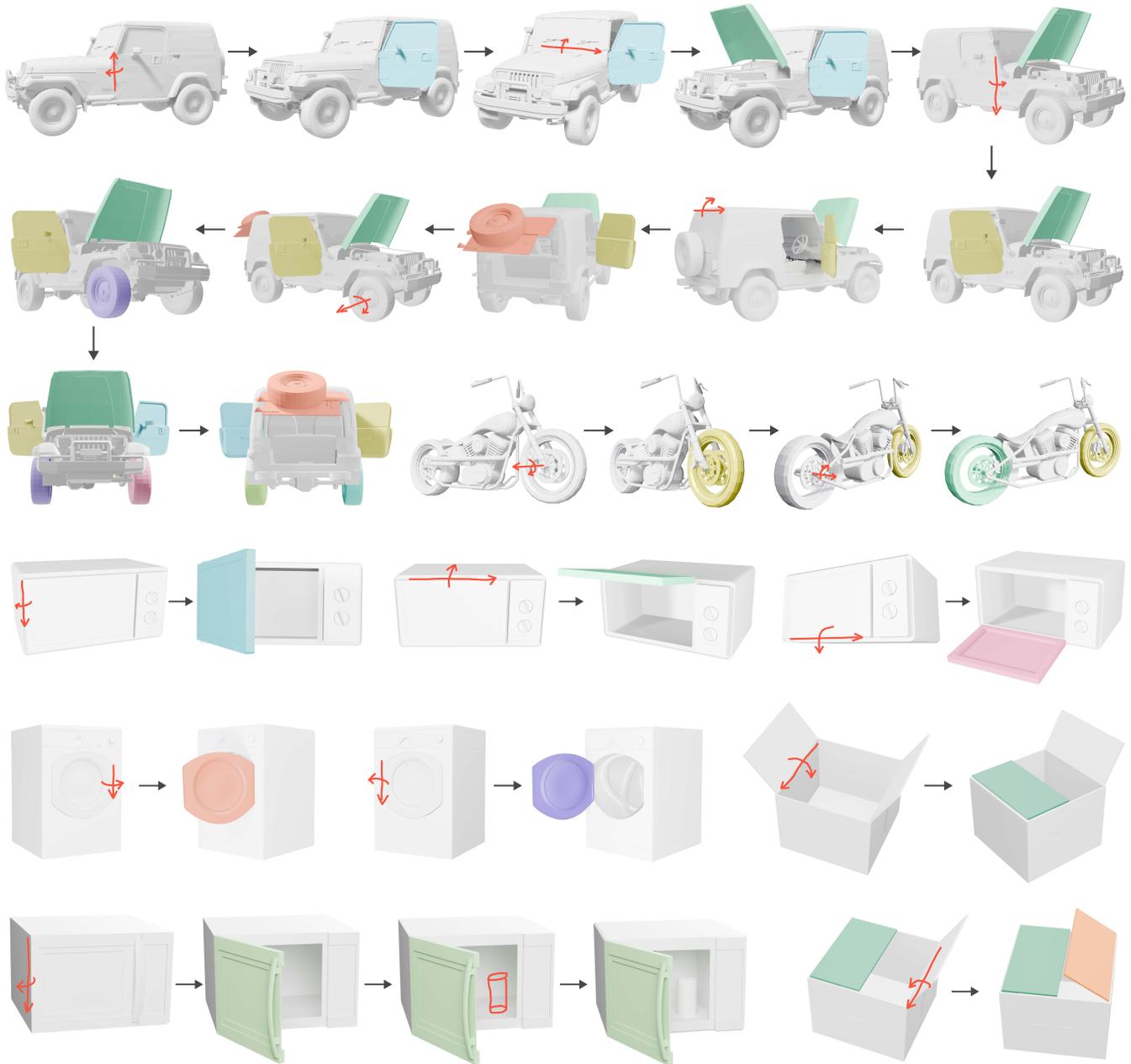}
    \caption{\textbf{Results gallery.}
    We show representative articulation modeling sessions using \name.
    For each example, user sketches are overlaid on the rendered shape under the chosen viewpoint, and the inferred movable parts are color-coded.
    The black arrow indicates the iterative modeling order across views/parts.}
    \label{fig:gallery}
\end{figure*}

\paragraph{Training data preparation.}
Given an articulated object in our dataset, we synthesize training samples through the following procedure.
First, we randomly select one movable part and sample a camera viewpoint from the upper hemisphere of a surrounding sphere, oriented to face the target part.
Second, we render the object under this viewpoint to obtain the part segmentation mask, depth map, and normal map, and generate the corresponding sketch cues as described above.
To emphasize local geometry around the articulated part, we constrain the camera distance so that the rendered maps focus on a close-up region.

Each synthesized training sample thus consists of a 2D sketch, the 2D movable-part mask, depth and normal maps capturing local geometry, and the ground-truth motion parameters.
By sampling multiple parts and viewpoints and varying the sketch perturbation parameters, we generate around 76K training samples in total.
We split the dataset into 70\%/20\%/10\% for training, testing, and validation, respectively.

\paragraph{Ethics and broader impacts}

\rev{We discuss the potential ethics and broader impacts of our dataset in the supplementary. A separate datasheet \cite{gebru2021datasheets,pushkarna2022data} can be found there.}

\section{Results and Evaluation}
\label{sec:results}
Using \name, we model articulated parts across diverse objects with varying geometric and kinematic complexity.
Representative results are shown in Figs.~\ref{fig:teaser} and \ref{fig:gallery}.
For complex models such as the car and the motorbike, users iteratively sketch from multiple viewpoints (often $>5$) to author articulations for many parts, ranging from bonnets to wheels.
Moreover, the same part can be articulated in different ways by simply providing different sketches (\eg car doors and microwave doors), demonstrating the high controllability enabled by our sketch-driven interface.
Thanks to the category-agnostic design of our predictor, \name generalizes well to uncommon objects rarely considered in existing benchmarks (\eg windmills and escape pods), producing plausible articulations without any class-specific adaptation.

To validate our approach, we conduct a user study (Sec.~\ref{subsec:user_eval}), comprehensive ablations (Sec.~\ref{subsec:ablation_arti} and Sec.~\ref{subsec:ablation_completion}), and quantitative comparisons against state-of-the-art articulation modeling baselines.
The supplementary video further demonstrates real-time articulation authoring with our system.

\paragraph{Implementation and runtime.}
Our articulation analysis network is implemented in PyTorch and trained on a single NVIDIA H200 GPU for one day.
We use Adam with a fixed learning rate of $10^{-4}$.
All inference experiments are conducted on a desktop equipped with a single NVIDIA A5000 GPU.

After loading a 3D object, PartField features require approximately $3$--$4$\,s to compute.
Although this step is relatively expensive, it can be overlapped with user navigation while selecting viewpoints and thus does not interrupt interaction in practice.
The articulation predictor runs in $\sim$30\,ms per forward pass.
The optional completion module takes around 30\,s \rev{per iteration step}, while other components (\eg articulation post-processing) run in negligible time. 
\rev{Simpler objects (\eg drawer) typically converge in just one iteration, complex objects (\eg wardrobe) that require 10-15 iterations take approximately $\sim$350s in total.}
Note that completion is optional and only performed upon user request.

\begin{figure*}[!t]
    \centering
    \begin{overpic}[width=0.99\textwidth]{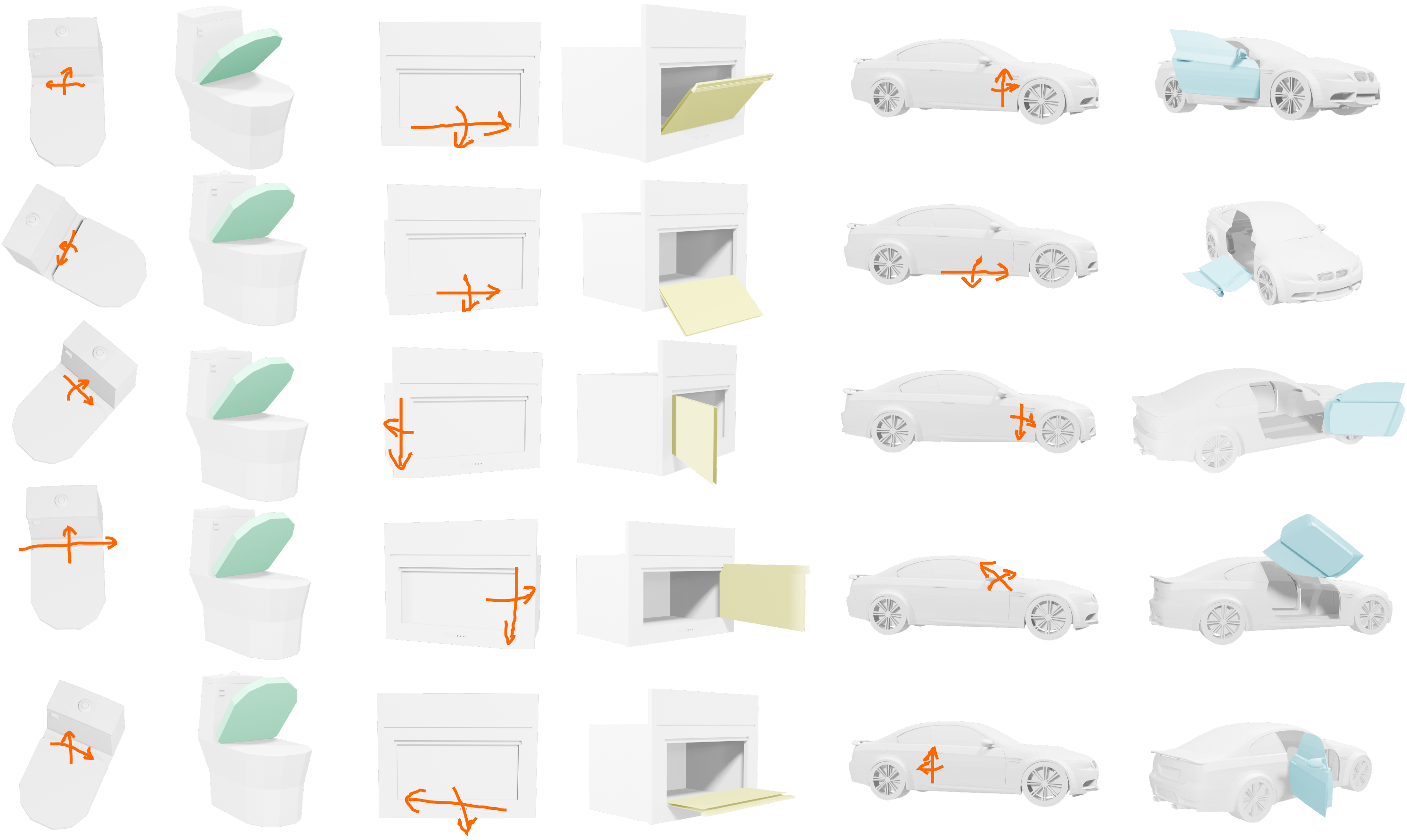} %
    \put(0, 57) {\small P$_1$}
    \put(0, 46) {\small P$_2$}
    \put(0, 36) {\small P$_3$}
    \put(0, 24) {\small P$_4$}
    \put(0, 11) {\small P$_5$}
    \end{overpic}
    \caption{\textbf{User gallery}. We asked 5 participants to model the articulation of three objects--a toilet, oven, and car. With a few coarse strokes, all users achieved their desired articulation.}
    \label{fig:user_study}
\end{figure*}
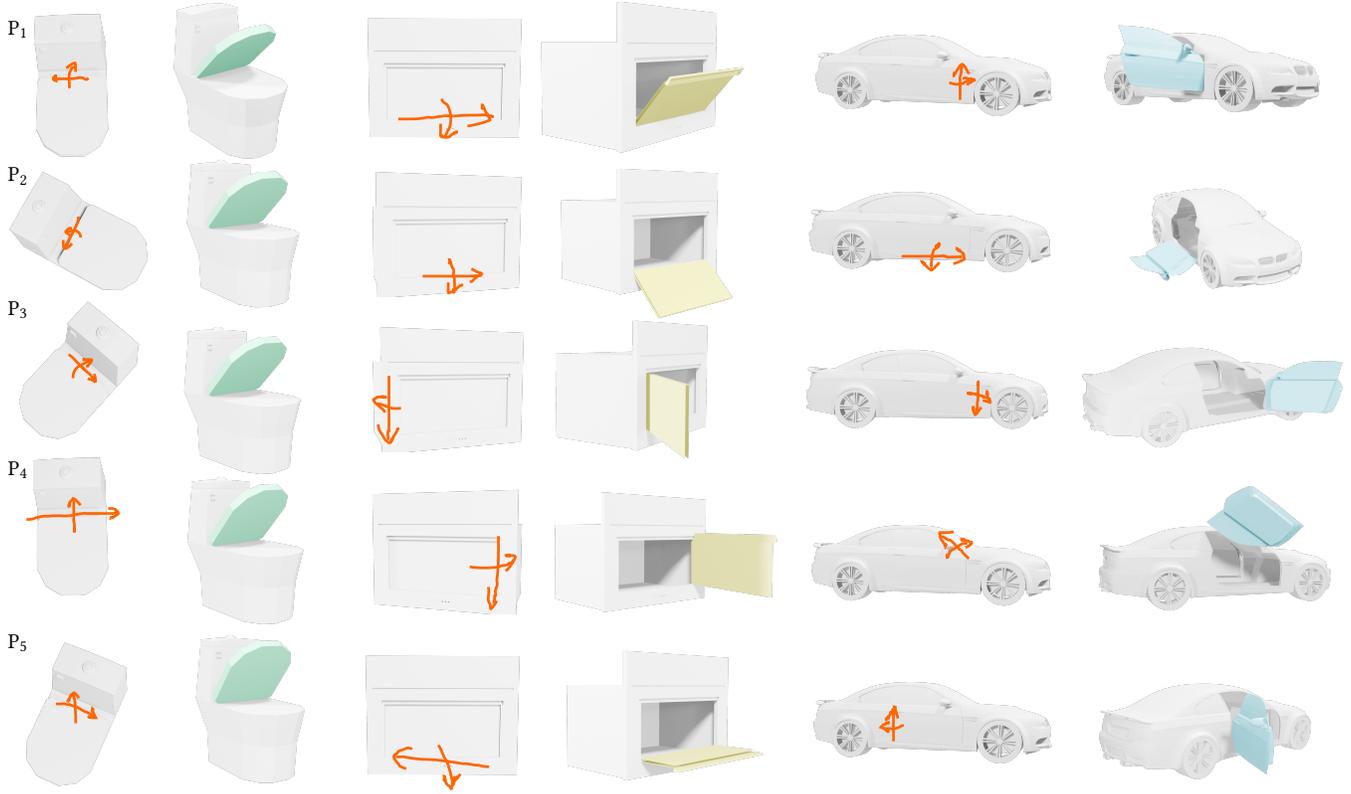

\subsection{User evaluation}
\label{subsec:user_eval}
We conducted a user study to assess the usability of our prototype system for sketch-based articulation modeling. 
Specifically, we invited five novice users and asked each of them to create articulations for the same three shapes: a toilet, an oven, and a car. 
For the toilet, participants were instructed to reproduce the intended articulation of the lid. 
For the oven and the car, we instead encouraged open-ended exploration, allowing participants to freely design alternative articulation mechanisms beyond conventional motions. 
All participants had varying levels of sketching and 3D modeling experience, but little to no prior exposure to articulation or animation tools. 
Before the study, each participant received a short tutorial followed by a brief hands-on practice session to get familiar with our system.

Figure~\ref{fig:user_study} summarizes the articulations produced by all participants. 
All users were able to quickly reproduce the toilet lid articulation, suggesting that our interface enables non-expert users to author basic articulated motions with minimal effort. 
In the exploratory tasks, participants frequently created diverse and sometimes unconventional articulation behaviors. 
For example, several participants designed different opening directions for the oven door and car door, and one participant (P2) intentionally produced a door motion that deviates from typical mechanical functionality. 
We interpret these variations as a desirable outcome of our sketch-based interaction, which supports expressive articulation authoring rather than restricting users to a small set of predefined behaviors.

The collected sketches in Fig.~\ref{fig:user_study} further indicate that our system can robustly interpret a wide range of stroke patterns that express similar articulation intent, highlighting the flexibility of freehand sketch inputs. 
Participants also provided suggestions for improving the workflow, such as incorporating automatic symmetry for repeated parts (e.g., wheels). 
Overall, the feedback was positive: users agreed that high-level sketch cues (\eg arrows and hinge strokes) can be reliably converted into coherent articulated structures.

In the post-study questionnaire, participants rated the correctness of sketch-to-motion translation at an average of 4.9/5, and rated the ease of ideation and exploration at an average of 4.7/5 (5-point Likert scale). 
Detailed participant feedback and qualitative comments are included in the supplementary material.

\begin{table*}[t]
  \centering
  \caption{\textbf{Quantitative Comparison.} We report statistics of the comparison between FreeArt3D~\cite{chen2025freeart3d}, Singapo~\cite{liu2024singapo}, and Ours. F-Score $\uparrow$, CD $\downarrow$, Joint-Axis-Err $\downarrow$, and Joint-Pivot-Err $\downarrow$ are reported. ``--'' indicates that the baseline method fails to handle the given category, or the metric is not available. ``Average-7'' refers to the average performance over the seven categories that all methods can handle. ``Average-10'' refers to the average performance over all ten categories.
}
\renewcommand{\arraystretch}{1.1} 
\resizebox{1.0\textwidth}{!}{
\begin{tabular}{c|c|*{10}{c}|c|c}
\toprule
\rowcolor{llightgray}
 & & Dishwasher & Laptop & Toilet & Microwave &
 Oven & Refrigerator &
 Door & Cabinet & Table &
 WashingMachine & Average-7 & Average-10 \\
\midrule

\multirow{5}[2]{*}{Ours}
& F-Score $\uparrow$ & \textbf{0.934} & \textbf{0.899} & \textbf{0.920} & 0.877 &
\textbf{0.932} & 0.908 & \textbf{0.864} &
\textbf{0.925} & \textbf{0.962} & \textbf{0.910} & \textbf{0.921} & \textbf{0.913} \\
& CD $\downarrow$ & \textbf{0.016} & \textbf{0.057} & \textbf{0.021} & 0.033 &
\textbf{0.027} & 0.024 & \textbf{0.034} &
\textbf{0.021} & \textbf{0.011} & \textbf{0.025} & \textbf{0.022} & \textbf{0.027} \\
& Joint-Axis-Err $\downarrow$ & \textbf{0.102} & \textbf{0.099} & \textbf{0.224} & 0.048 &
\textbf{0.130} & 0.218 & \textbf{0.246} &
0.297 & 0.456 & \textbf{0.378} & \textbf{0.233} & \textbf{0.220} \\
& Joint-Pivot-Err $\downarrow$ & \textbf{0.086} & \textbf{0.143} & \textbf{0.366} & \textbf{0.209} &
\textbf{0.157} & \textbf{0.065} & \textbf{0.281} &
- & - & \textbf{0.143} & \textbf{0.132} & \textbf{0.181} \\
\midrule

\multirow{5}[2]{*}{\shortstack[c]{FreeArt3D\\\textit{(Siggraph Asia 2025)}}}
& F-Score $\uparrow$ & 0.915 & 0.864 & 0.854 & \textbf{0.890} &
0.883 & \textbf{0.943} & 0.778 &
0.914 & 0.924 & 0.854 & 0.903 & 0.882 \\
& CD $\downarrow$ & 0.024 & 0.065 & 0.029 & \textbf{0.031} &
0.046 & \textbf{0.022} & 0.042 &
0.026 & 0.015 & 0.031 & 0.028 & 0.033 \\
& Joint-Axis-Err $\downarrow$ & 0.248 & 0.176 & 0.524 & \textbf{0.032} &
0.430 & 0.365 & 0.319 &
0.591 & 1.082 & 0.978 & 0.532 & 0.475 \\
& Joint-Pivot-Err $\downarrow$ & 0.131 & 0.160 & 0.514 & 0.283 &
0.204 & 0.095 & 0.360 &
- & - & 0.353 & 0.285 & 0.263 \\
\midrule

\multirow{5}[2]{*}{\shortstack[c]{Singapo\\\textit{(ICLR 2025)}}}
& F-Score $\uparrow$ & 0.766 & - & - & 0.726 &
0.866 & 0.844 & - &
0.760 & 0.800 & 0.833 & 0.799 & - \\
& CD $\downarrow$ & 0.053 & - & - & 0.045 &
0.035 & 0.031 & - &
0.037 & 0.041 & 0.030 & 0.039 & - \\
& Joint-Axis-Err $\downarrow$ & 0.293 & - & - & 0.231 &
0.339 & \textbf{0.155} & - &
\textbf{0.179} & \textbf{0.102} & 0.578 & 0.268 & - \\
& Joint-Pivot-Err $\downarrow$ & 0.134 & - & - & 0.248 &
0.302 & 0.115 & - &
- & - & 0.229 & 0.205 & - \\
\bottomrule
\end{tabular}
}
\label{tab:in_domain_comparison}
\end{table*}

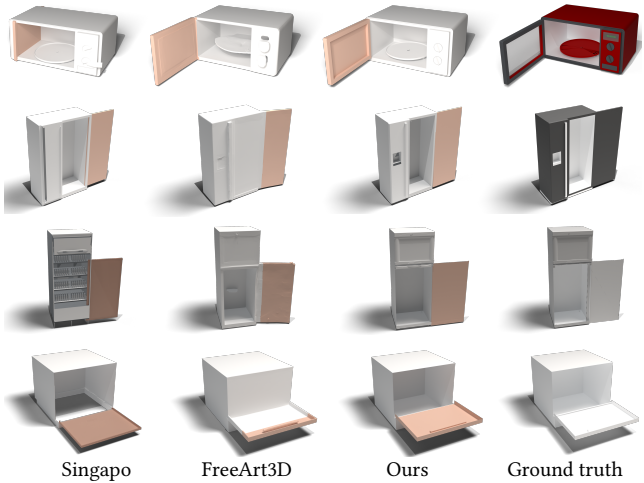
\begin{figure}[!t]
    \centering
    \begin{overpic}[width=0.99\linewidth]{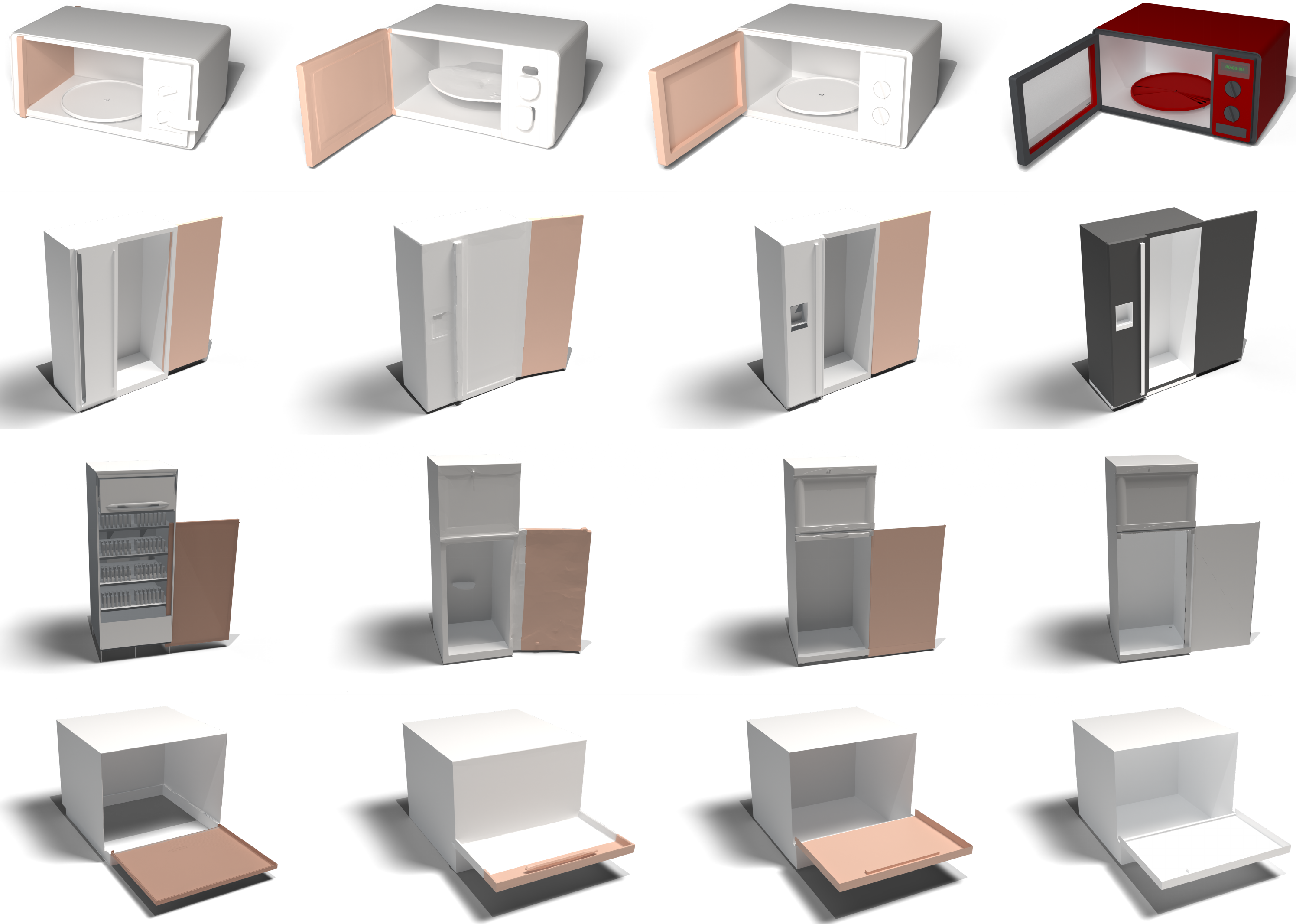} %
    \put(9,-3) {\small Singapo}
    \put(31,-3) {\small FreeArt3D}
    \put(60,-3) {\small Ours}
    \put(79,-3) {\small Ground truth}
    \end{overpic}
    \caption{\textbf{Visual comparison.}
    We show four representative examples comparing Singapo, FreeArt3D, and our method against the ground truth.}
    \label{fig:visual_comparison}
\end{figure}

\begin{table*}[t]
  \centering
  \caption{\textbf{Out-of-Distribution Quantitative Comparison}. 
  We report results of FreeArt3D~\cite{chen2025freeart3d} and Ours in terms of 
  F-Score $\uparrow$, CD $\downarrow$, Joint-Axis-Err $\downarrow$, and Joint-Pivot-Err $\downarrow$. 
  ``--'' indicates that the baseline method fails to handle the given category. 
  ``Average-5'' denotes the average over the five categories that both methods can handle, 
  and ``Average-10'' denotes the average over all ten categories.}
  \renewcommand{\arraystretch}{1.1}
  \resizebox{1.0\textwidth}{!}{
  \begin{tabular}{c|c|*{10}{c}|c|c}
    \toprule
    \rowcolor{llightgray}
    & & Car & Lighter & Box & Suitcase & Window$^{\text{R}}$
    & Skateboard & Fan & Bike & Motorbike & Windmill
    & Average-5 & Average-10 \\
    \midrule
    \multirow{4}[2]{*}{Ours}
    & F-Score $\uparrow$
      & \textbf{0.955} & \textbf{0.923} & \textbf{0.927} & \textbf{0.900} & \textbf{0.891}
      & \textbf{0.943} & \textbf{0.911} & \textbf{0.877} & \textbf{0.925} & \textbf{0.903}
      & \textbf{0.919} & \textbf{0.916} \\
    & CD $\downarrow$
      & \textbf{0.014} & \textbf{0.022} & \textbf{0.026} & \textbf{0.031} & \textbf{0.028}
      & \textbf{0.032} & \textbf{0.054} & \textbf{0.028} & \textbf{0.034} & \textbf{0.015}
      & \textbf{0.024} & \textbf{0.028} \\
    & Joint-Axis-Err $\downarrow$
      & \textbf{0.102} & \textbf{0.067} & \textbf{0.066} & \textbf{0.109} & \textbf{0.096}
      & \textbf{0.087} & \textbf{0.099} & \textbf{0.102} & \textbf{0.078} & \textbf{0.096}
      & \textbf{0.088} & \textbf{0.090} \\
    & Joint-Pivot-Err $\downarrow$
      & \textbf{0.034} & \textbf{0.036} & \textbf{0.061} & \textbf{0.083} & \textbf{0.079}
      & \textbf{0.025} & \textbf{0.065} & \textbf{0.099} & \textbf{0.048} & \textbf{0.091}
      & \textbf{0.059} & \textbf{0.062} \\
    \midrule
    \multirow{4}[2]{*}{\shortstack[c]{FreeArt3D\\\textit{(Siggraph Asia 2025)}}}
    & F-Score $\uparrow$
      & 0.928 & 0.883 & 0.903 & 0.814 & 0.862
      & -- & -- & -- & -- & --
      & 0.878 & -- \\
    & CD $\downarrow$
      & 0.018 & 0.036 & 0.029 & 0.040 & 0.039
      & -- & -- & -- & -- & --
      & 0.032 & -- \\
    & Joint-Axis-Err $\downarrow$
      & 1.302 & 0.123 & 0.094 & 0.121 & 0.101
      & -- & -- & -- & -- & --
      & 0.348 & -- \\
    & Joint-Pivot-Err $\downarrow$
      & 0.237 & 0.072 & 0.098 & 0.100 & 0.104
      & -- & -- & -- & -- & --
      & 0.122 & -- \\
    \bottomrule
  \end{tabular}
  }
  \label{tab:ood_comparison}
\end{table*}

\subsection{Comparison}
\label{subsec:comp}

We compare our articulation prediction approach against state-of-the-art methods. In what follows, we elaborate on details.

\paragraph{Competitors.}
Two representative state-of-the-art methods for articulated object modeling are chosen---FreeArt3D \cite{chen2025freeart3d} and Singapo \cite{liu2024singapo}. 
FreeArt3D requires six input images, each captured at a distinct joint configuration. These configurations are uniformly sampled between the rest pose and the maximum articulation, following their experimental \rev{settings}. 
In contrast, Singapo operates under a single-view setting. It constructs articulated objects by retrieving and assembling semantic parts from the PartNet-Mobility dataset~\cite{xiang2020sapien}. 

\paragraph{Experiment setting and competitors.}
For a given test example, we construct the desired inputs for each method.
For our method, the inputs are the 3D object and the synthesized sketch.
For FreeArt3D, the inputs are six rendered images of the 3D object (same viewpoint as ours). Each image captures the evolving articulation status from the rest pose to the maximum articulation.
As for Singapo, the input is a rendered rest pose image of the 3D object (same viewpoint as ours).
We use their published codes with the default parameters and the released networks with default weights, following their execution steps to run the experiment.

\paragraph{Dataset.}
To facilitate the comparison, we choose 10 object categories (\ie Dishwasher, Laptop, Toilet, Microwave, Oven, Refrigerator, Door, Cabinet, Table, and Washing Machine) from our \dname dataset.
For each category, we randomly select 50 shapes, yielding a total of 500 test instances.
Compared to using a small test subset from PartNet-Mobility, our dataset is constructed at a similar semantic level but includes a larger number of evaluation instances. 
While the object categories remain unchanged, the data exhibits moderate distributional variations, resulting in minor yet meaningful differences in object structure and visual composition, such as the presence of additional elements on tabletops.

Regarding category coverage, FreeArt3D supports all ten evalution categories, whereas Singapo is applicable to only seven of them (see Tab.~\ref{tab:in_domain_comparison}).

\paragraph{Metrics.} 
We evaluate each predicted articulated shape, including the static body, movable parts, and estimated joint parameters, using four metrics: Chamfer Distance (CD)~\cite{fan2017point}, F-Score~\cite{wang2018pixel2mesh}, joint axis direction error, and joint pivot error.
Specifically, CD is computed by sampling 100k surface points, and F-Score is evaluated with a threshold of 0.05.
We adopt the same definition of joint axis direction error and the joint pivot error as in \cite{le2024articulate, chen2025freeart3d}.

Following~\cite{chen2025freeart3d}, for each shape, we sample six joint states and generate the corresponding articulated meshes using the predicted joint parameters. All metrics are computed for each joint state and averaged across states. Prior to evaluation, the generated meshes are rigidly aligned to the ground-truth meshes using~\cite{liu2024one}.

\paragraph{Evaluation.}
Table~\ref{tab:in_domain_comparison} reports quantitative comparisons.
Overall, \name achieves the best average performance across all four metrics over the evaluated categories.
On the seven categories supported by all baselines, our method improves the F-score by 2.0\% and 15.3\% over FreeArt3D and Singapo, respectively, while reducing CD by 21.3\% and 43.6\%.
For motion estimation, \name further yields substantial gains in articulation accuracy, improving the joint axis error by 56.2\% / 13.1\% and the joint pivot error by 53.7\% / 35.6\% compared to FreeArt3D and Singapo.
On the remaining three categories where only FreeArt3D and our method are applicable, \name consistently outperforms FreeArt3D on all metrics.
These results demonstrate the effectiveness of our sketch-driven articulation modeling.

Qualitative comparisons are shown in Fig.~\ref{fig:visual_comparison}.
FreeArt3D often exhibits two characteristic failure modes.
First, although the target movable part is articulated, a duplicate of the part at its initial pose is frequently generated (\eg the microwave and refrigerator doors).
Second, despite leveraging Trellis, its original feed-forward generation tends to alter the input geometry and fails to preserve either the static structure or the revealed interior after articulation.
For instance, the rotary knob and tray of the microwave are significantly distorted, and spurious internal artifacts appear inside the wardrobe.
This comparison also highlights the importance of our articulation-aware adaptation of Trellis, which enables constrained inpainting that preserves the given geometry under articulated motion.
Singapo relies on retrieval and part assembly. 
While the overall configuration may roughly match the target, the geometric fidelity is limited.
In contrast, \name accurately recovers movable parts and articulation parameters while better preserving the original geometry.

\paragraph{Generalization to out-of-domain shapes.}
We also test our method and compare it with the most recent baseline FreeArt3D on a set of categories not covered by PartNet-Mobility or our training set.
As shown in Tab.~\ref{tab:ood_comparison}, 
\rev{the ten OOD categories are: Car, Lighter, Box, Suitcase, Rotational Windows (denoted as Window$^R$), Skateboard, Fan, Bike, Motorbike, and Windmill. Among these, the last five categories}
contain continuous rotations that are \emph{not yet} supported by FreeArt3D, to which our approach conveniently extends.
Meanwhile, on the categories both methods can handle, our approach produces results more accurate than FreeArt3D, especially on the \rev{car} joint axis dimension.
For FreeArt3D, its 3D generative model-driven articulation optimization may not capture the continuous movement of car parts (\eg doors) well, because of the overall complexity of a car shape and the relatively small portion of parts like doors.
On the other hand, our localized sketch-based inference does not rely on recognizing the car category but robustly infers the local motions of movable parts generalizable across categories, like doors, hoods, and wheels. 

This contrast shows that our sketch-based articulation modeling with a focus on generalization indeed excels in handling novel object categories.
Note that Singapo is excluded from this comparison, because as a retrieval-based method, Singapo uses a part database from PartNet-Mobility and thus cannot handle such novel categories, as well as that Singapo does not account for continuous rotations either.

\subsection{Ablation Study - Articulation Analysis}
\label{subsec:ablation_arti}

\begin{figure}[!t]
    \centering
    \includegraphics[width=0.99\linewidth]{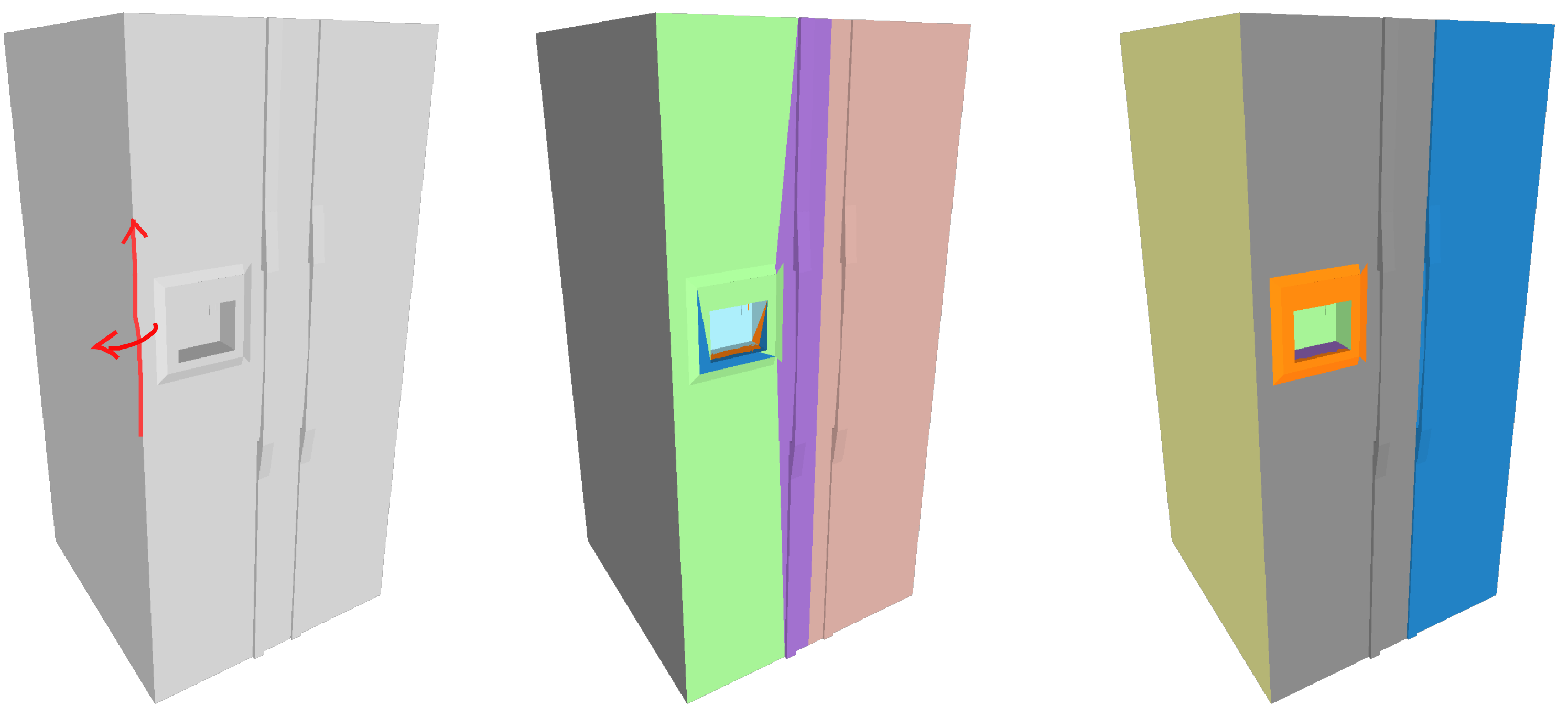}
    \caption{\textbf{Part segmentation.}
    (a) Given a user sketch, we localize the target movable part using PartField features guided by the predicted part cues.
    (b) A k-means baseline yields infeasible segments with cross-part boundaries due to its flat clustering (see the purple segment and the ice outlet).
    (c) Our hierarchical strategy produces more plausible, part-consistent segments by organizing neighboring clusters in a tree structure.
    }
    \label{fig:abl_segmenation}
\end{figure}

\paragraph{Hierarchical segmentation.}
As described in Sec.~\ref{subsec:segmentation}, we leverage PartField features and propose a hierarchical clustering strategy to select the 3D part(s) that best match the predicted part cues (\ie the 2D segmentation map) by matching 3D clusters to the 2D cues.
To validate its effectiveness, we compare our hierarchical segmentation with a k-means baseline.
For k-means, we apply the same cluster-to-cue matching procedure by selecting the cluster that maximizes IoU with the predicted cues.
We empirically tune the number of clusters and report the best-performing setting (\ie $k=20$).

Figure~\ref{fig:abl_segmenation} presents a qualitative comparison.
We observe that k-means often produces infeasible part segments, either fragmenting a single semantic part into multiple clusters or merging geometrically adjacent but functionally distinct components.
This behavior is expected, as k-means imposes a flat partition of the feature space and lacks the ability to capture the hierarchical and part-consistent structure of articulated components.
In contrast, our hierarchical segmentation consistently identifies the correct movable part(s) and yields cleaner boundaries.

For highly complex shapes with dense interior structures, we occasionally observe imperfect selections (\eg including interior components when extracting an outer surface part).
In such cases, our interface allows users to easily correct the segmentation by clicking to add or remove candidate parts, enabling lightweight error correction without disrupting the workflow.

\begin{figure}[!t]
    \centering
    \includegraphics[width=0.87\linewidth]{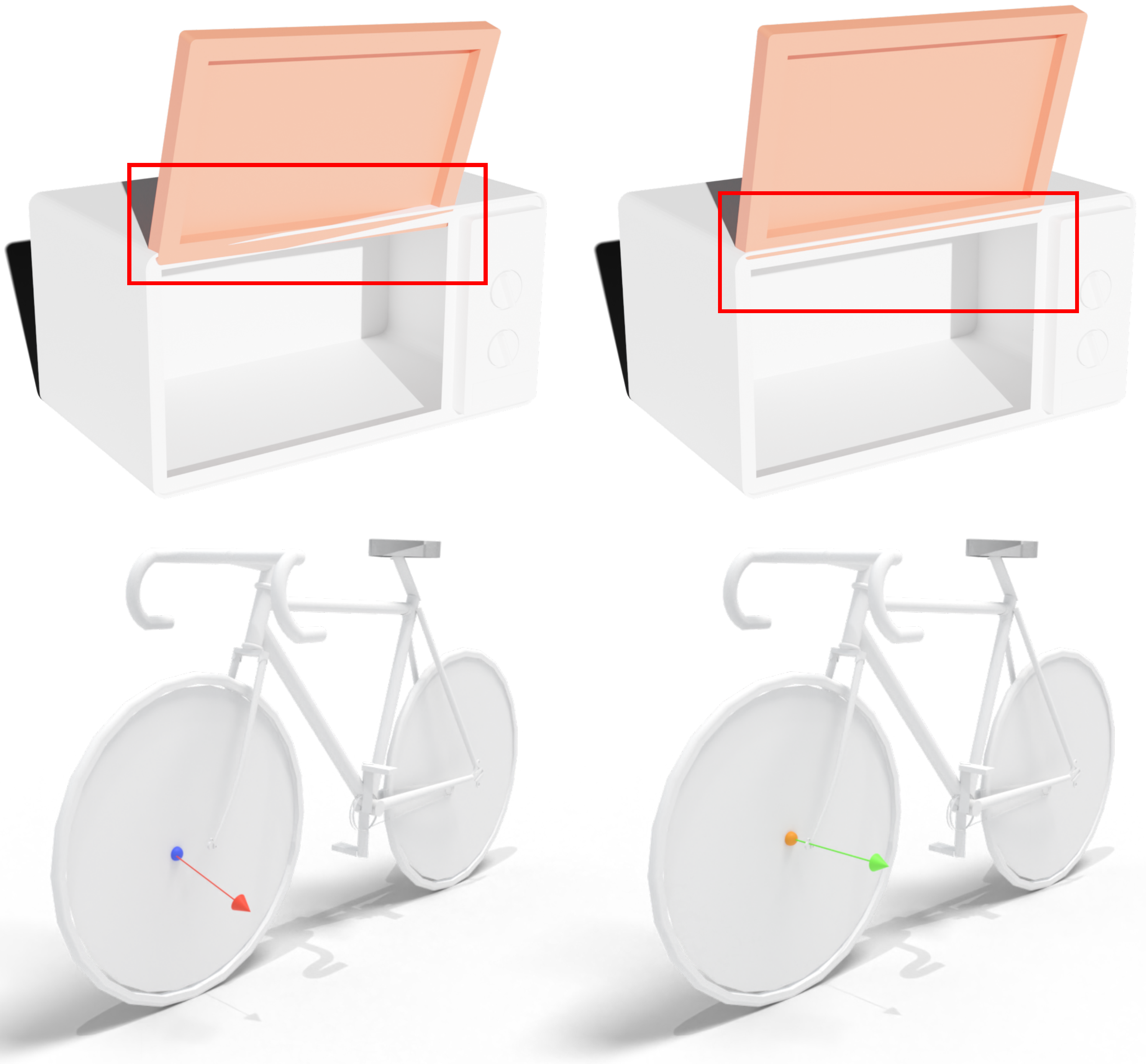}
    \caption{\textbf{Geometric snapping.}
    We compare articulation predictions \textit{w/o} (left) and \textit{w/} (right) geometric snapping.
    \textbf{Top:} microwave door articulation. Without snapping, the predicted axis/pivot slightly deviates from the hinge geometry, leading to misaligned opening. Snapping anchors the parameters to local geometric cues and yields a plausible hinge motion.
    \textbf{Bottom:} bicycle front-wheel articulation. Snapping refines the predicted rotation axis and pivot point to better align with the wheel structure.}
    \label{fig:abl_snapping}
\end{figure}

\paragraph{Geometric snapping.}
To evaluate the effectiveness of our geometric snapping procedure (Sec.~\ref{subsec:post-processing}), we compare predictions with and without this post-processing step.
Figure~\ref{fig:abl_snapping}(left) shows that, although the network outputs exhibit small numerical errors, slight misalignments of the motion axis and pivot can still be clearly observed when visualized in 3D.
Such subtle deviations from local geometric structures (e.g., hinges or part boundaries) often lead to visually implausible or mechanically inconsistent articulations.

As shown in Fig.~\ref{fig:abl_snapping}(right), our geometric snapping reliably corrects these errors by anchoring the predicted parameters to salient local geometry, providing a strong geometric reference.
This post-processing step thus improves the physical plausibility and visual fidelity of the resulting articulation.

\begin{table}[!t]
    \centering
    \small
    \caption{\textbf{Ablation Study.} We compare Trellis \cite{xiang2025structured} (a), FreeArt3D \cite{chen2025freeart3d} (b) and our variants, in the following dimensions: \textbf{Struct.}: structure preservation; \textbf{Comp.}: completeness and integrity; \textbf{Bound.}: boundary confinement; \textbf{Valid.}: kinematic validity.}
    \begin{tabular*}{\columnwidth}{l@{\extracolsep{\fill}}cccc}
        \toprule
        \textbf{Config.} & \textbf{Struct.} & \textbf{Comp.} & \textbf{Bound.} & \textbf{Valid.} \\
        \midrule
        (a) Trellis baseline & - & \checkmark & - & - \\
        (b) Disk Norm. & - & \checkmark & - & - \\
        (c) + Masked Comp. & \checkmark & - & \checkmark & - \\
        (d) + Iteration & \checkmark & \checkmark & \checkmark & - \\
        \midrule
        \textbf{(e) + Kinematic Ref.} & \textbf{\checkmark} & \textbf{\checkmark} & \textbf{\checkmark} & \textbf{\checkmark} \\
        \bottomrule
    \end{tabular*}
    \label{tab:ablation}
\end{table}

\subsection{Ablation Study - Interior Completion}
\label{subsec:ablation_completion}

We validate our design for interior completion through a component-wise ablation study, summarized in Table \ref{tab:ablation}.

\paragraph{Structure Preservation}
Both FreeArt3D (\ie disk normalization, Tab. \ref{tab:ablation}(b)) and our method build upon the Trellis generative model for 3D shape completion \rev{Tab. \ref{tab:ablation}}(a). 
While FreeArt3D introduces an auxiliary disk to normalize scale for image-to-3D generation, it remains an unconstrained process, as it is not required to align with any pre-existing 3D geometry. 
However, for our structure-preserving interior completion, we face a stricter requirement: the completion must respect the static geometry and fit together tightly. 
Thus the lack of explicit 3D constraints in previous methods becomes a critical limitation. 

To address this, we introduced Masked Subspace Completion (Tab. \ref{tab:ablation}(c)), which locks the latent state to the known shell, explicitly enforcing the presence of the original structure.
Meanwhile, Negative Masking addresses the over-generation risk. 
By effectively constraining growth within logical bounds, it prevents the geometry from spilling into invalid regions, such as artificially thickening the drawer base. 

\begin{figure}[!t]
    \centering
    \includegraphics[width=0.85\linewidth]{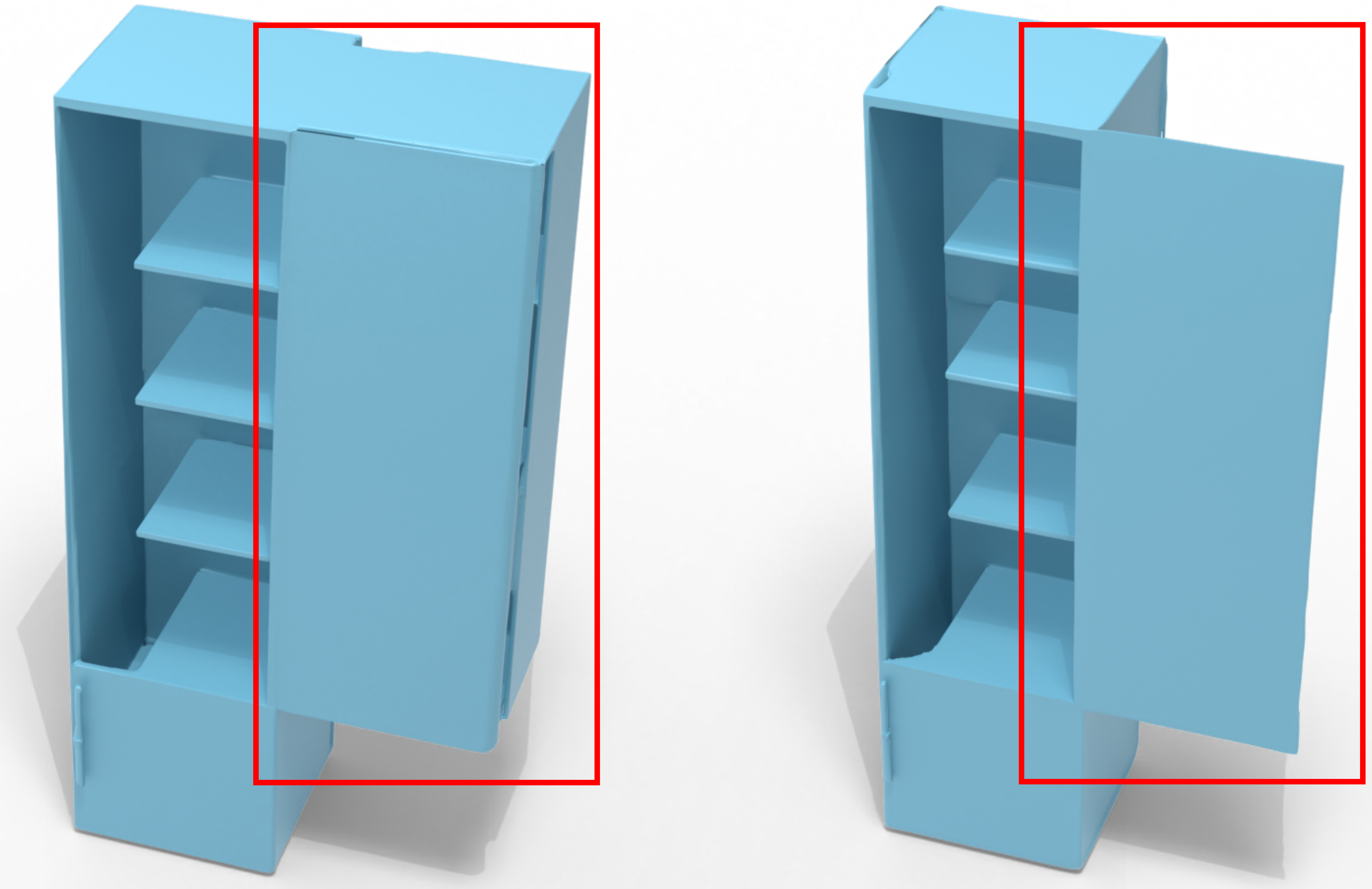}
    \caption{\textbf{Ablation on masked completion} for structure preservation. Masked completion of interior structures enables both the preservation of given static structures and the avoidance of extra erroneous content in the void. On the left, without masked completion, the cabinet has a drifted size and extra shape for the opened door. On the right, with masked completion, a clean cabinet of proper size and shape has been created, although the interior structure is not exactly complete and needs iterations to fix (Fig.~\ref{fig:iteration}).}
    \label{fig:abl_mask_completion}
\end{figure}

Fig.~\ref{fig:abl_mask_completion} shows such a comparison with and without the masked completion for structure preservation. 
When completing without mask control, the generative model tends to produce extra content and change the reference shape globally.
With the mask control, we can obtain a shape with precise alignment with the given input, and no extra undesirable content is hallucinated.

\begin{figure}[!t]
    \centering
    \includegraphics[width=\linewidth]{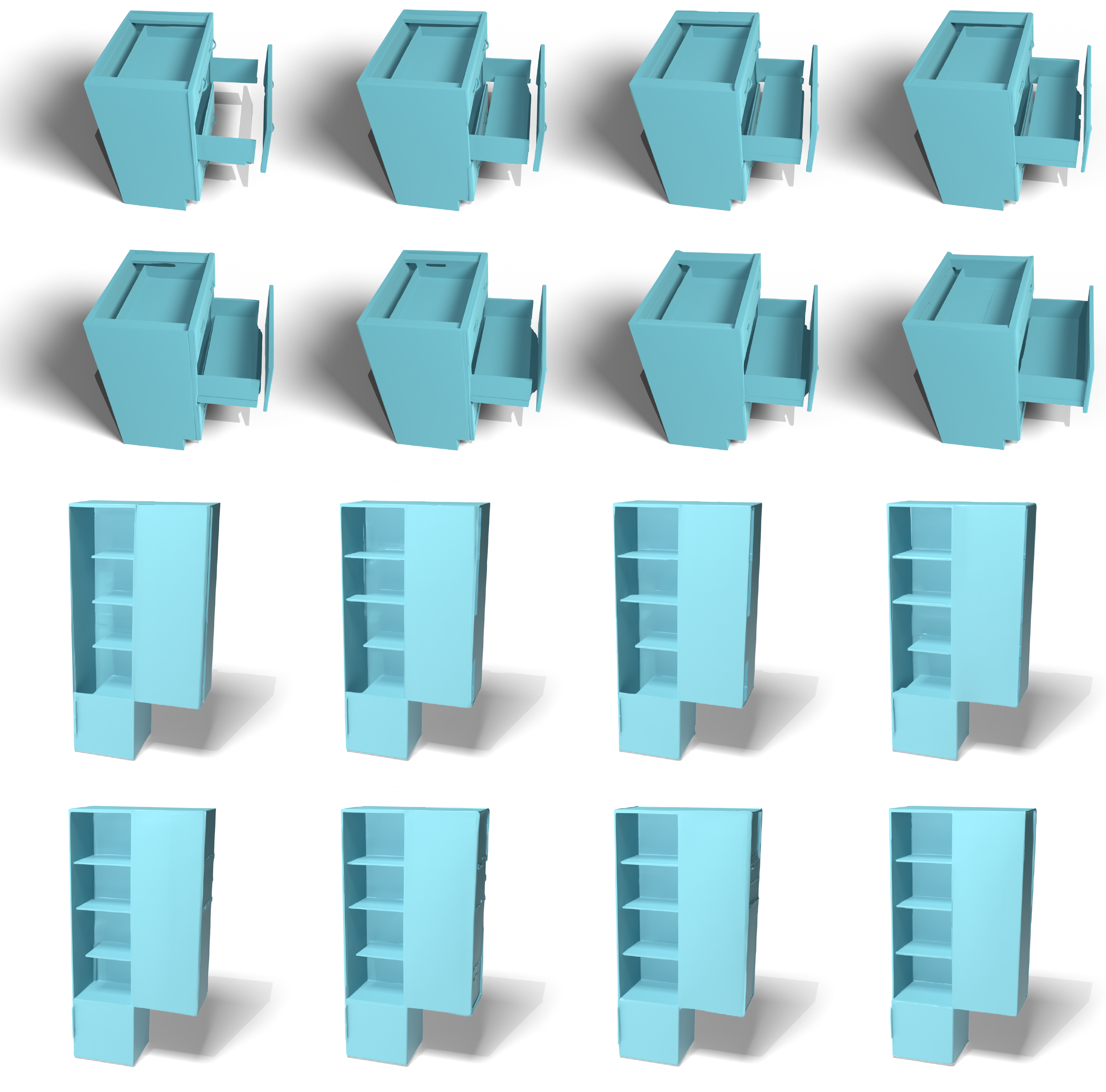}
    \caption{\textbf{Ablation on iterated completion}. For each test case, the iteration starts from the top left and continues to the down right. With each intermediate shape one more pass of iterative generation is applied, gradually completing the interior structure. The difference between the initial completion and the final completion highlights the limited capacity of existing generative models for interior structure completion, and the necessity of our iterative scheme. }
    \label{fig:iteration}
\end{figure}

\paragraph{Iteration}

As discussed in Sec. \ref{sec:flow_completion}, single-pass generation often fails to fully bridge the structural gap, yielding only partial components. To resolve this problem, we introduced Iterative Refinement (Tab. \ref{tab:ablation} (d)). 
As shown in Fig.~\ref{fig:iteration}, multi-pass generation allows the complex interior geometry to progressively grow and consolidate, finally achieving completeness.

\begin{figure}[!t]
    \centering
    \begin{overpic}[width=\linewidth]{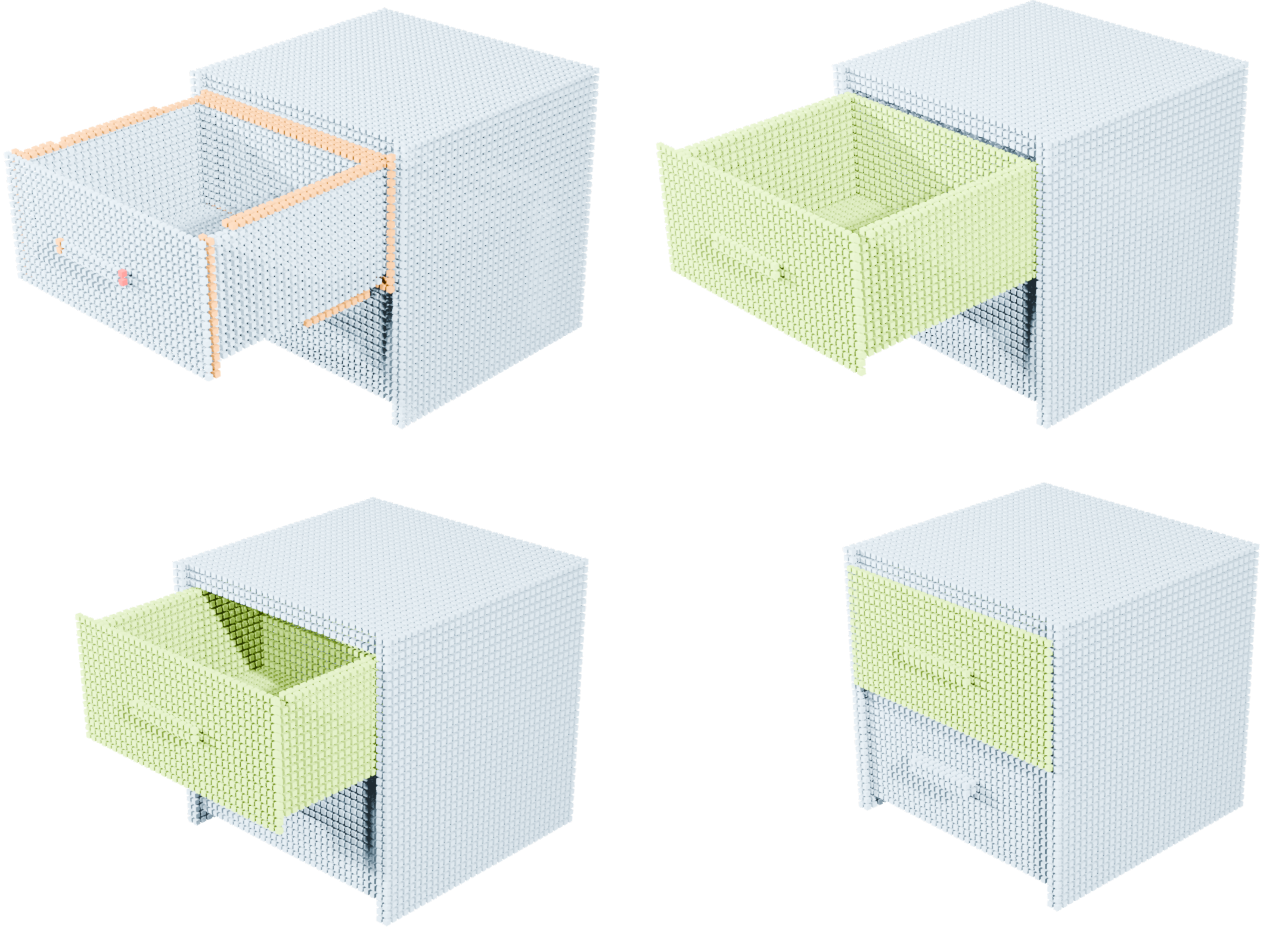} %
    \put(0, 41) {\small (a)}
    \put(54, 41) {\small (b)}
    \put(0, 2) {\small (c)}
    \put(54, 2) {\small (d)}
    \end{overpic}
    \caption{\textbf{Ablation on Kinematic Refinement.} (a) Without Kinematic Refinement, the generation suffers from missing voxels (red) and produces excess voxels that collide with the base part (\rev{orange}). (b-d) With our Kinematic Refinement applied, the model can be articulated smoothly and without collision.}
    \label{fig:abl_kinematic_refinement}
\end{figure}

\paragraph{Kinematic Refinement}
While the asset generated by iterative masked completion is visually complete, it is not yet articulation-ready. 
The synthesized moving part may be overly wide, causing collisions with the base during motion, or too short to fulfill the intended range. 
Our final Kinematic Refinement (Tab. \ref{tab:ablation} (e)) addresses this by performing collision detection and length calibration directly in the voxel space, as shown in Fig.~\ref{fig:abl_kinematic_refinement}. 

Furthermore, since Trellis employs FlexiCubes \cite{shen2023flexicubes} for mesh extraction, generating the moving and static parts together often leads to their fusion as a single component. By explicitly decoupling the generation process, we naturally prevent such artifacts and obtain separated mesh components. This facilitates the automatic application of motion parameters and the export of a valid URDF, marking the transition from a static visual mesh to a functional model compatible with physics simulation.

\subsection{Limitation and Future Work}
\label{subsec:Limi_and_discuss}

\begin{figure}[!t]
    \centering
    \includegraphics[width=0.8\linewidth]{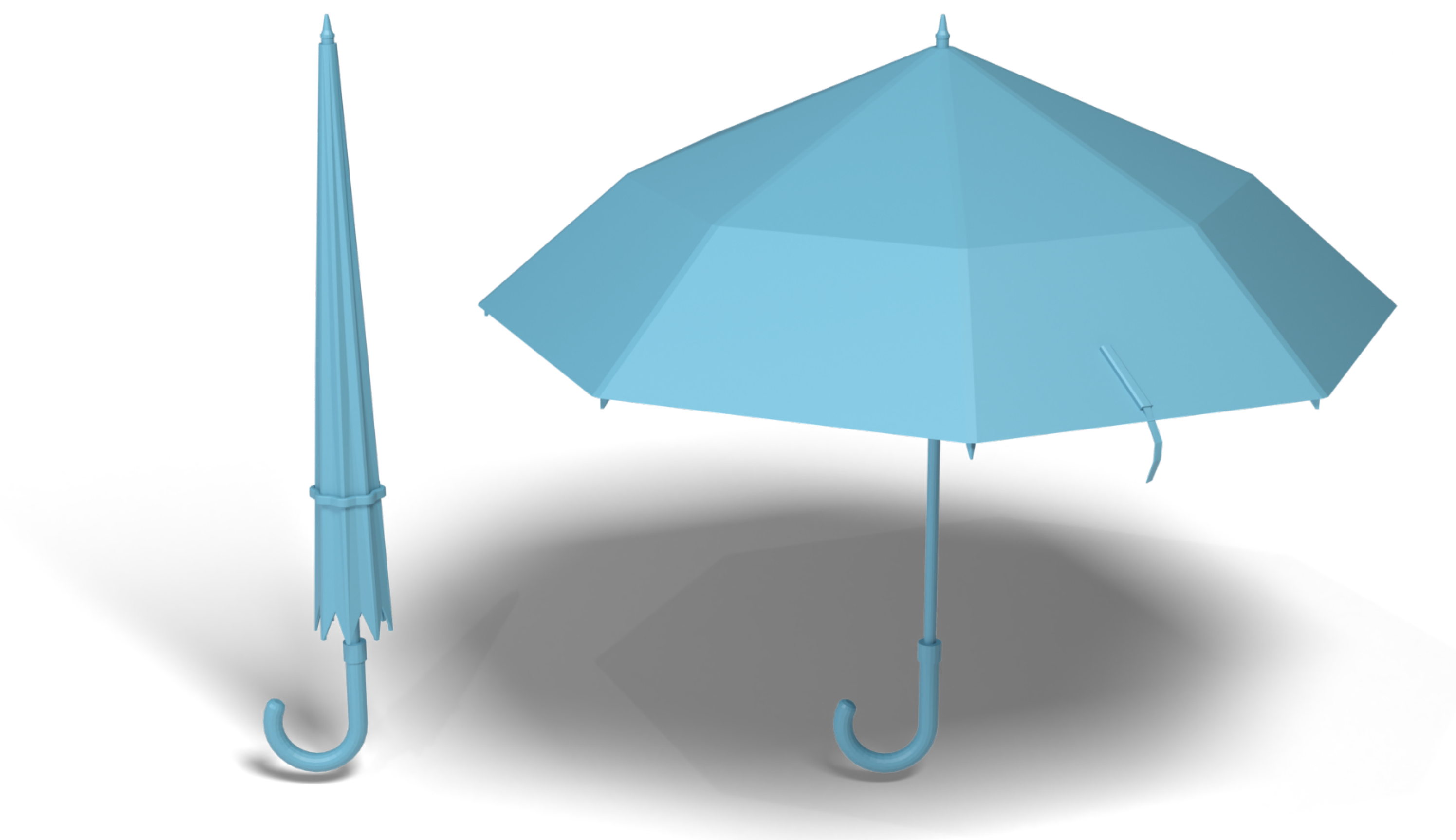}
    \caption{\textbf{Limitation.} Opening an umbrella requires complex, coupled articulation across many parts, which cannot be captured by our current single-joint motion model.}
    \label{fig:failure_case}
\end{figure}

\paragraph{Limited articulation types.}
A limitation of our current approach is that it primarily targets simple articulations that can be parameterized by a single motion model, \eg rotation or translation.
As a canonical failure case, Fig.~\ref{fig:failure_case} shows the opening of an umbrella, which cannot be adequately described by a single set of articulation parameters.
This motion involves tightly coupled interactions among multiple components: a translational slide along the shaft triggers coordinated rotations of many struts and a global non-rigid deformation of the canopy.
Since \name is designed to predict part-level rigid articulations with relatively constrained motion models, it cannot currently capture such complex, multi-part, and coupled mechanisms.
Extending sketch-based articulation modeling to handle these compound articulations remains an interesting direction for future work.

\rev{\paragraph{Input mesh topology.}
Built upon Partfield features, we segment the movable part from the input mesh model. However, when the mesh tesselation is messy, \name does not perform mesh cutting or face splitting to recover semantic articulation boundaries, which exceeds Partfield's capacity.
A more advanced semantic-aware mesh segmentation model or light-weight user interactions can help overcome this limitation.
}

\paragraph{Joint geometry and articulation modeling.}
Another limitation stems from the fact that many existing 3D meshes are not constructed with articulation constraints in mind~\cite{pan2015flow,yu2022piecewise,zhang2025variational}.
As a result, the input geometry often lacks explicit joint structures or geometric cues that indicate motion axes and contact regions.
For instance, a car door in a typical mesh model may be optimized for surface appearance and smoothness, yet omit physically meaningful hinge geometry, which makes accurate axis alignment and motion specification inherently ambiguous.
This observation motivates an exciting future direction: \emph{articulation-aware modeling}, where 3D assets are not merely visual shells but functional geometries with topologies and joint structures that explicitly support physical interaction and realistic motion.

\paragraph{Towards unified sketch-based shape and articulation modeling.}
Sketch-based shape modeling has been studied for decades, while articulation modeling has recently gained increasing attention.
In this work, we take a first step towards bridging these two domains by enabling sketch-driven articulation modeling on top of an existing 3D shape.
A more ambitious and ultimately desirable goal, however, is a unified system that jointly models \emph{both} shape geometry and articulation from scratch, as envisioned in Fig.~\ref{fig:motivation}.
In such a setting, geometry and articulation can mutually constrain and reinforce each other---\eg articulation-aware geometry supports physically meaningful joint structures, while articulation intent guides the functional design of shape parts.
We believe this joint modeling paradigm would be a key ingredient for creating application-ready 3D assets that are not only visually plausible but also kinematically functional.

\section{Conclusion}
We introduced \name, the first sketch-based articulation modeling system for CAD objects.
Given a 3D shape and lightweight user sketches drawn from a chosen viewpoint, \name automatically identifies movable parts and predicts their articulation parameters, enabling iterative and highly controllable articulation authoring.
Technically, our method combines sketch-guided articulation analysis with a geometry-aware part extraction strategy, and further supports articulation-aware, sketch-controllable shape inpainting by adapting generative models to complete missing internal structures while respecting inferred motion constraints.
Extensive experiments and user evaluations demonstrate that our method achieves strong controllability and generalization across diverse objects beyond existing articulation datasets.

\begin{acks}
The authors would like to thank the reviewers for their valuable suggestions, Haocheng Yuan for preparing the visual figures, Chen Wang for preparing the pipeline figures, Wenhan Xue for preparing the comparison figures, and Junyang Cheng for data filtering. CL was supported by a gift from Adobe.
\end{acks}

\bibliographystyle{ACM-Reference-Format}
\bibliography{main}

\clearpage
\section*{Supplemental Material}
\appendix

\setcounter{table}{0}
\renewcommand{\thetable}{A\arabic{table}}
\setcounter{figure}{0}
\renewcommand{\thefigure}{A\arabic{figure}}

In this supplementary material, \rev{we provide additional details on the user study protocol and raw responses, the full text prompt used when interacting with the VLM, and the datasheet containing the full information on the new \dname dataset.}

\section{User Study}

We provide the complete data of the 5 participants evaluating our \name system here.

\noindent\textbf{Participants.}
The participants include two postgraduate students, two master students, and one undergraduate student, all from the computer science discipline. \rev{They were randomly recruited from the university campus.} We also take note of their background experiences with sketch drawing, 3D modeling, and articulated or animation-related modeling.

\noindent\textbf{Procedure.}
Each evaluation consists of three parts:

\paragraph{a. Training session.}
We communicate with the participant the basic idea of \name, and demonstrate how sketched arrows are used to specify object articulation and motion. In particular, we introduce the two supported motion types: translation, specified by a single directional arrow, and rotation, specified by a hinge (axis) line together with a direction arrow. The participant is then allowed to freely try the tool through casual modeling and exploration.

\paragraph{b. Modeling session.}
Participants are asked to perform articulation modeling tasks. For the toilet example, participants are required to create an articulated shape similar to the given example. For the oven and car examples, participants are given flexibility to either reproduce similar articulation or freely explore their own articulated designs. During the modeling session, we only answer questions related to UI usage.

\paragraph{c. Discussion session.}
We ask feedback from the participant by going through the following four questions:

\begin{itemize}
    \item Q1: Is the articulation modeling process easy to conceive and explore? (Rating: 0--5)
    \item Q2: Is the sketching intuitive and properly translated into the intended motion types and parameters? (Rating: 0--5)
    \item Q3: Does the resulting motion and articulation quality meet your expectations? (Rating: 0--5)
    \item Q4: Overall rating, with comments on positive and negative aspects.
\end{itemize}

\subsection{User evaluation data divided by participant}

\emph{Participant \#1.}

\textbf{Background.}
\begin{itemize}
    \item \textit{Sketching:} Knows the basic concept of sketching, but has no formal training in drawing.
    \item \textit{3D modeling:} Has watched demo videos before, but has never tried 3D modeling personally.
    \item \textit{Articulation / animation:} Familiar with everyday articulated objects. Has no prior experience with animation tools or animation production.

\end{itemize}

\textbf{Session duration.}
\begin{itemize}
    \item \textit{Training session:} 15 minutes.
    \item \textit{Modeling session:}
    task~1: 5 minutes; task~2: 4 minutes; task~3: 11 minutes.
\end{itemize}

\textbf{Ratings.}
\begin{itemize}
    \item Q1: 5, Q2: 5, Q3: 5, Q4: 4.5
\end{itemize}

\textbf{Comments.}

\textit{Negative.}
\begin{itemize}
    \item Mouse-based sketching is less convenient and less precise compared to using a tablet with pen input.
    \item Simple sketch strokes are prone to deviation during drawing, indicating the need for stroke pre-processing to improve sketching convenience.
\end{itemize}

\textit{Positive.}
\begin{itemize}
    \item The sketch-based interaction is easy to learn and straightforward to use.
    \item The quality of the resulting articulated models is high.
\end{itemize}

\vspace{2mm}
\emph{Participant \#2.}

\textbf{Background.}
\begin{itemize}
    \item \textit{Sketching:} Has taken a course in product sketch design.
    \item \textit{3D modeling:} Has limited experience with 3D modeling in Blender.
    \item \textit{Articulation / animation:} Has limited experience with animation in Blender.
\end{itemize}

\textbf{Session duration.}
\begin{itemize}
    \item \textit{Training session:} 20 minutes.
    \item \textit{Modeling session:}
    task~1: 4 minutes; task~2: 8 minutes; task~3: 25 minutes.
\end{itemize}

\textbf{Ratings.}
\begin{itemize}
    \item Q1: 4, Q2: 5, Q3: 5, Q4: 4.5
\end{itemize}

\textbf{Comments.}

\textit{Negative.}
\begin{itemize}
    \item For symmetric objects such as cars, it takes a lot of effort to define the same motion for repeated parts (e.g., wheels). I hope the system can support symmetric articulation so that I do not need to draw the same sketches multiple times.
    \item Since wheels usually move in continuous and fairly fixed ways, it would be useful if the system could automatically generate articulation for wheels. This would allow me to spend more time on designing new and unique parts.
\end{itemize}

\textit{Positive.}
\begin{itemize}
    \item I tried different design alternatives in the car example. Some door opening mechanisms I designed are not realistic for real-world use (e.g., doors opening downward), but this made the system feel interesting to explore.
    \item Being able to create motion directly from geometric shapes is nice and still gives me freedom to try different designs.
    \item Overall, I am satisfied with how the motions turn out.
\end{itemize}

\vspace{2mm}
\emph{Participant \#3.}

\textbf{Background.}
\begin{itemize}
    \item \textit{Sketching:} Has basic experience with sketching.
    \item \textit{3D modeling:} No prior experience.
    \item \textit{Articulation / animation:} No prior experience.
\end{itemize}

\textbf{Session duration.}
\begin{itemize}
    \item \textit{Training session:} 20 minutes.
    \item \textit{Modeling session:}
    task~1: 5 minutes; task~2: 7 minutes; task~3: 12 minutes.
\end{itemize}

\textbf{Ratings.}
\begin{itemize}
    \item Q1: 5; Q2: 5; Q3: 5; Q4: 5
\end{itemize}

\textbf{Comments.}

\textit{Negative.}
\begin{itemize}
    \item The user interface could look better, and some buttons are not very comfortable to use.
\end{itemize}

\textit{Positive.}
\begin{itemize}
    \item I could reproduce the example results by following the simple sketching steps shown in the tutorial.
    \item The system is easy to learn and easy to use.
\end{itemize}

\vspace{2mm}
\emph{Participant \#4.}

\textbf{Background.}
\begin{itemize}
    \item \textit{Sketching:} Has some experience with basic sketching.
    \item \textit{3D modeling:} Has basic experience with 3D modeling tools.
    \item \textit{Articulation / animation:} No prior experience.
\end{itemize}

\textbf{Session duration.}
\begin{itemize}
    \item \textit{Training session:} 25 minutes.
    \item \textit{Modeling session:}
    task~1: 6 minutes; task~2: 10 minutes; task~3: 18 minutes.
\end{itemize}

\textbf{Ratings.}
\begin{itemize}
    \item Q1: 5; Q2: 4.5; Q3: 5; Q4: 5
\end{itemize}

\textbf{Comments.}

\textit{Negative.}
\begin{itemize}
    \item I did not encounter any obvious negative issues.
\end{itemize}

\textit{Positive.}
\begin{itemize}
    \item I like the animation buttons, as they allow me to see my model move.
\end{itemize}

\vspace{2mm}
\emph{Participant \#5.}

\textbf{Background.}
\begin{itemize}
    \item \textit{Sketching:} Has basic experience with freehand sketching and drawing simple shapes.
    \item \textit{3D modeling:} Has limited experience with entry-level 3D modeling software.
    \item \textit{Articulation / animation:} No prior experience.
\end{itemize}

\textbf{Session duration.}
\begin{itemize}
    \item \textit{Training session:} 20 minutes.
    \item \textit{Modeling session:}
    task~1: 5 minutes; task~2: 6 minutes; task~3: 9 minutes.
\end{itemize}

\textbf{Ratings.}
\begin{itemize}
    \item Q1: 4.5
    \item Q2: 5
    \item Q3: 4.5
    \item Q4: 5
\end{itemize}

\textbf{Comments.}

\textit{Negative.}
\begin{itemize}
    \item No significant negative issues were encountered during the session.
\end{itemize}

\textit{Positive.}
\begin{itemize}
    \item The system interface is intuitive and easy to understand after a short training period.
\end{itemize}

\section{Nano Banana Prompt}

\begin{minipage}{\columnwidth}
\begin{lstlisting}[label={lst:nano_banana_prompt}]
Input: A surface-only articulated object in an open state rendered from a user-selected 3D viewpoint, together with a user-drawn sketch indicating the geometry to be generated <(*@$\mathbf{I}_{in}, \mathbf{S}$@*)>

Task_Prompt:
You are tasked with completing geometries from user sketches. The input includes an articulated object represented only by its exterior surfaces, where the interior geometry is intentionally unspecified, and a sketch drawn by the user to define the desired interior structure.

Rules you must follow:
- Follow the sketch exactly. The sketch defines the geometry to be generated. You must not modify its shape, reinterpret it, or invent alternatives.
- Do not include the sketch lines in the output. They serve only as a boundary or guideline.
- Preserve the structure of the original object. The generated geometry must seamlessly attach to the existing surfaces.
- Allow only minimal adjustments (e.g., scaling, alignment) to ensure physical plausibility and proper attachment. These adjustments are for fitting, not for changing the shape.
- Never deviate from the sketch's geometry. The shape provided by the user must be realized exactly.
- The generated geometry must have the same material and appearance as the original object, so that the completion blends naturally with the visible surfaces.

Example:
If the sketch outlines a triangular drawer as the interior structure, you must generate a triangular drawer that attaches correctly to the object. You may adjust its size or alignment so it fits, but you must not alter its triangular shape, and the drawer's material must match the original object.

Output:
A completed object image rendered from the same user-selected 3D viewpoint, with the sketch-defined geometry integrated into the object <(*@$\mathbf{I}_{comp}$@*)>.
\end{lstlisting}
\end{minipage}

\begin{table*}[t]
\centering
\caption{\textbf{Category distribution of \dname.} Instance counts for each object category in the dataset.}

\setlength{\tabcolsep}{10pt}
\renewcommand{\arraystretch}{1.05}
\resizebox{0.8\textwidth}{!}{
\begin{tabular}{l r | l r | l r}
\toprule
\rowcolor{llightgray}
\textbf{Category} & \textbf{\#Instances} &
\textbf{Category} & \textbf{\#Instances} &
\textbf{Category} & \textbf{\#Instances} \\
\midrule
bottle          & 903 & oven            & 85 & lighter         & 37 \\
table           & 512 & refrigerator    & 81 & swing           & 36 \\
eyeglasses      & 243 & washing machine & 78 & valve           & 36 \\
cabinet         & 230 & windmill        & 78 & stapler         & 33 \\
monitor         & 230 & screwdriver     & 70 & globe           & 29 \\
skateboard      & 201 & lamp            & 68 & seesaw          & 23 \\
faucet          & 162 & bike            & 63 & folding chair   & 21 \\
toilet          & 161 & kettle          & 61 & swivel chair    & 21 \\
laptop          & 148 & clock           & 58 & rocking chair   & 19 \\
fan             & 139 & scissors        & 54 & knife           & 17 \\
bucket          & 124 & pen             & 52 & window          & 14 \\
handcart        & 121 & dishwasher      & 50 & carton          & 8  \\
motorbike       & 107 & microwave       & 50 & watch           & 6  \\
car             & 101 & plier           & 45 & & \\
door            & 96  & excavator       & 39 & & \\
\midrule
\rowcolor{llightgray}
\multicolumn{5}{r}{\textbf{Total}} & \textbf{4710} \\
\bottomrule
\end{tabular}
}
\label{tab:dataset_category_stats}
\end{table*}

\section{Dataset Details}
We provide the detailed per-category instance numbers in Tab.~\ref{tab:dataset_category_stats}. 
\rev{A separate Datasheet can be found in the \texttt{datasheet.pdf} file.}

\rev{
\paragraph{Ethics and broader impacts}
To promote transparency and responsible AI development, we document the \dname dataset following established frameworks \cite{gebru2021datasheets,pushkarna2022data}. Our dataset is constructed from public research datasets \cite{wang2019shape2motion, wang2025partnext}, together with controlled procedural generation \cite{joshi2505procedural}. This design follows common scholarly data-sourcing practices and does not rely on web scraping.

To improve dataset provenance and traceability, we provide a metadata mapping file that links each of the 4,710 shapes to its immediate upstream source and identifier, facilitating attribution to original creators within the research lineage \cite{longpre2024data}.

Regarding broader impacts, our system emphasizes human-in-the-loop interaction through sketch-based control. This design preserves user agency in articulation modeling and offers a more interpretable alternative to fully automated generation, which may help reduce unintended or opaque outcomes.
}

\end{document}